\documentclass{article}

\usepackage[preprint]{neurips_2025}

\usepackage[utf8]{inputenc} %
\usepackage[T1]{fontenc}    %
\usepackage{url}            %
\usepackage{booktabs}       %
\usepackage{amsfonts}       %
\usepackage{nicefrac}       %
\usepackage{microtype}      %
\usepackage{xcolor}         %

\usepackage[frozencache=true]{minted}
\usepackage{caption}
\usepackage{algorithm}
\usepackage[noend]{algpseudocode}

\usepackage{booktabs}
\usepackage{longtable}
\usepackage{threeparttable}
\usepackage{multirow}

\usepackage{amsmath}
\usepackage{amssymb}
\usepackage{mathtools}
\usepackage{amsthm}

\usepackage{graphicx}
\usepackage{subcaption}
\usepackage{tikz}           %
\usetikzlibrary{patterns,patterns.meta,positioning,backgrounds,calc}

\definecolor{darkGreen}{rgb}{0.2,0.5,0.2}
\definecolor{mydarkblue}{rgb}{0,0.08,0.45}
\usepackage[colorlinks,citecolor=mydarkblue,urlcolor=mydarkblue,linkcolor=mydarkblue]{hyperref}
\usepackage[capitalize,noabbrev]{cleveref}

\usepackage{enumitem}
\usepackage{etoc}

\usepackage{tcolorbox}
\tcbuselibrary{breakable}

\definecolor{metadatacolor}{RGB}{52, 73, 94}      
\definecolor{taskcolor}{RGB}{142, 68, 173}        
\definecolor{calibrationcolor}{RGB}{39, 174, 96}  
\definecolor{instructioncolor}{RGB}{211, 84, 0}   

\usepackage{amsmath,amsfonts,bm}

\def\eqref#1{equation~\ref{#1}}

\def\1{\bm{1}}

\DeclareMathAlphabet{\mathsfit}{\encodingdefault}{\sfdefault}{m}{sl}
\SetMathAlphabet{\mathsfit}{bold}{\encodingdefault}{\sfdefault}{bx}{n}

\usepackage{todonotes}
\usepackage[normalem]{ulem}

\makeatletter
\newcommand*\iftodonotes{\if@todonotes@disabled\expandafter\@secondoftwo\else\expandafter\@firstoftwo\fi}  %
\makeatother

\newcommand{\affilgithub}{\raisebox{-1.5pt}{\includegraphics[height=0.9em]{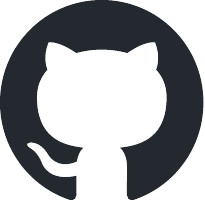}}\,}

\title{BRIDGE: Predicting Human Task Completion Time From Model Performance}

\author{
  Fengyuan Liu$^{\text{*} \ 1,2}$\,\,
  Jay Gala$^{\text{*} \ 1,2}$\,\,
  Nilaksh$^{1,3}$\,\,
  Dzmitry Bahdanau$^{1,2,4,5}$\,\,
  Siva Reddy$^{1,2,5,6}$\,\,
  Hugo Larochelle$^{1}$
  \vspace{0.4em} \\
  {\small
  $^1$Mila - Quebec AI Institute \,\,$^2$McGill University \,\,$^3$Polytechnique Montréal \,\,$^4$Periodic Labs  \\
  \small
  $^5$Canada CIFAR AI Chair\,\,$^6$ServiceNow Research \vspace{0.1em}
  \vspace{0.5em}\\

  \scriptsize
  $^*$Equal Contribution \\\ \ {Correspondence:  \texttt{\{fengyuan.liu,jay.gala\}@mila.quebec}}
  }
}

\begin{document}

\maketitle


\begin{abstract}
Evaluating the real-world capabilities of AI systems requires grounding benchmark performance in human-interpretable measures of task difficulty. Existing approaches that rely on direct human task completion time annotations are costly, noisy, and difficult to scale across benchmarks. In this work, we propose BRIDGE, a unified psychometric framework that learns a latent difficulty scale from model responses and anchors it to human task completion time. Using a two-parameter logistic Item Response Theory model, we jointly estimate latent task difficulty and model capability from model performance data across multiple benchmarks. We demonstrate that latent task difficulty varies linearly with the logarithm of human completion time, allowing human task completion time to be inferred for new benchmarks from model performance alone. Leveraging this alignment, we forecast frontier model capabilities in terms of human task length and independently reproduce METR’s exponential scaling results, with the 50\% solvable task horizon doubling approximately every 6 months.

\vspace{3mm}
{\small\sffamily \textbf{Code Repository:} \affilgithub\href{https://github.com/McGill-NLP/BRIDGE}{\textbf{\texttt{McGill-NLP/BRIDGE}}}}

\end{abstract}


\section{Introduction}

As Artificial Intelligence (AI) systems are increasingly deployed in open-ended, real-world settings, their capabilities are typically reported via benchmark scores and aggregate metrics. However, such scores are difficult to interpret as measures of \emph{task difficulty} or \emph{real-world effort}: improvements may reflect gains on short, routine instances while leaving longer, multi-step tasks largely unchanged, and comparable score changes can correspond to very different shifts in practical capability. Consequently, benchmark performance alone provides limited guidance about what AI systems can reliably do or how quickly practical capability is improving. A more actionable evaluation should express performance on human-aligned scales, most directly, the time a human would require to complete a task.

A prominent attempt to express AI capability in human terms is by \citet{kwa2025measuringaiabilitycomplete} at METR, which measures performance in terms of the length of tasks AI agents can complete and reports exponential growth, with task-length horizons doubling roughly every 7 months. While compelling, this paradigm depends on human task completion time annotations from people with relevant expertise. These annotations are expensive collect and are difficult to extend consistently across diverse benchmarks. As model capabilities continue to scale, relying on new human studies to anchor each benchmark becomes increasingly impractical, creating a growing gap between benchmark-centric evaluation and human-centric notions of difficulty.

\begin{figure}[t]
    \centerline{\includegraphics[width=0.6\columnwidth]{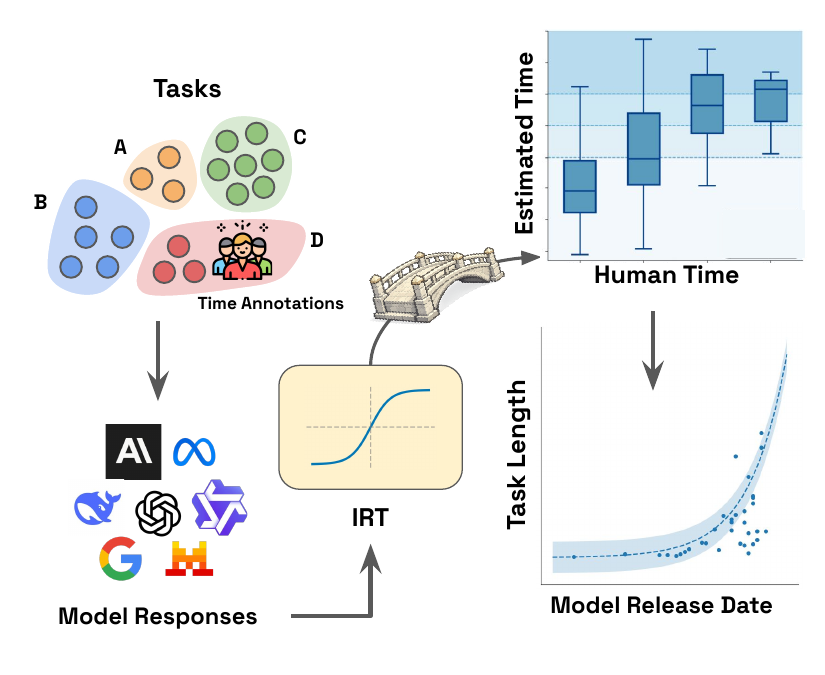}}
    \caption{
        Overview of BRIDGE. Model responses across different benchmarks (clustered by colors) are used to fit a two-parameter logistic Item Response Theory (2PL IRT) model, estimating latent task difficulty and model capability. Calibrating latent difficulty against tasks with known human task completion times yields accurate predictions of human task duration for new benchmarks. We leverage this alignment to forecast frontier model capabilities in terms of human task length even in the absence of human task duration annotations.
    }
    \label{fig:pipeline}
\end{figure}

In this work, we introduce BRIDGE,\footnote{BRIDGE refers to \textbf{B}enchmark \textbf{R}esponse \textbf{I}nferred \textbf{D}ifficulty \textbf{G}rounded in \textbf{E}lapsed human time} a unified psychometric framework (illustrated in \Cref{fig:pipeline}) that addresses this gap by aligning task difficulty for humans, measured by task completion time, with task difficulty for models, measured by benchmark performance. 

Item Response Theory (IRT) \citep{baker2001basics}, and in particular the two-parameter logistic (2PL) model, has been adopted to analyze large language model (LLM) performance and benchmarks \citep{Lalor2019LearningLP, Rodriguez2021EvaluationEA, hofmann2025fluidlanguagemodelbenchmarking}. We adopt a 2PL IRT model to jointly estimate the latent difficulty of individual tasks and the capability of individual models using a binarized indicator (success vs. failure) as performance data across multiple benchmarks.

Our central empirical finding is that IRT-inferred latent task difficulty closely tracks human completion time: across benchmarks with annotations, latent difficulty varies approximately linearly with the logarithm of human task duration. This relationship anchors the latent scale to a human-interpretable time axis and enables prediction of human task completion time from model performance alone. After calibrating the scale using existing human annotations (e.g., METR), we can map IRT difficulty values for new benchmarks onto this time scale without conducting new human studies. We show that for newly introduced tasks (e.g., SWE-bench \cite{jimenez2024swebench} and Cybench \cite{zhang2025cybench}), our predicted completion times align well with both available human annotations and qualitative expectations, indicating that the IRT-based latent scale effectively transfers difficulty estimates between humans and models.

Building on this alignment, we use BRIDGE to forecast frontier model capabilities in terms of human task-length horizons using only model performance logs, without requiring human task-time annotations. We estimate that solvable task length continues to grow exponentially, with a doubling time of approximately 6 months. These trends are consistent with the findings of \citet{kwa2025measuringaiabilitycomplete}, providing independent validation that model-centric data can recover human-centric task length forecasting.


\section{Background and Motivation}

\subsection{Measuring capability through human task completion time}
\label{subsec:metrbkgd}

Estimating task completion time provides a meaningful bridge between benchmarks and real-world applications, enabling AI capability to be expressed in human-interpretable units and to forecast progress over different time horizons \citep{t-AGI}. METR \citep{kwa2025measuringaiabilitycomplete} operationalized this idea through the ``50\%-task-completion time horizon,'' defined as the duration of tasks that an AI system can complete with 50\% probability. Using this metric, they showed that the frontier task-length horizon has grown exponentially in recent years. Their analysis relied on human time annotations for 170 tasks, collected from annotators with relevant domain knowledge but no task-specific context.

Although the METR framework provides a valuable human-centered scale, its reliance on direct time annotations tightly couples each benchmark to a bespoke human study. This coupling makes it difficult to propagate a calibrated notion of difficulty across benchmarks with different formats, domains, or task distributions. METR’s subsequent work partially addresses cross-domain transfer by estimating human completion time under alternative modeling assumptions; however, this approach still depends on benchmark-specific heuristics and strong priors about task structure \citep{how-does-time-horizon-vary-across-domains}, which become increasingly fragile as benchmarks grow more heterogeneous and open-ended (e.g., GDPval \citep{patwardhan2025gdpvalevaluatingaimodel}). More broadly, repeated human studies do not scale with the pace at which new benchmarks and task variants are introduced. These limitations raise a natural question: can we predict human task completion time without conducting new human studies for each benchmark? Our work addresses this question by introducing a latent, model-derived difficulty scale that is explicitly calibrated to human completion time, enabling scalable prediction of human task duration for new benchmarks.

\subsection{Item response theory for LLM benchmarking}
\label{subsec:irtbkgd}
Psychometric methods originally developed for educational testing, particularly IRT \cite{baker2001basics}, provide a principled framework for modeling latent ability\footnote{We use \emph{ability} to denote the latent trait parameter $\theta$ estimated via IRT and \emph{capability} to refer to the underlying construct it approximates.} of individuals and latent difficulty of tasks. Given a set of tasks $T = \{t_1, \dots, t_{|T|}\}$, each task $t_i$ is characterized by a discrimination parameter $a_i \geq 0$, which measures how strongly the task differentiates models with different capabilities and a difficulty parameter $b$.
For the set of models $M = \{m_1, \dots, m_{|M|}\}$, each model $m_j$ has its ability estimate $\theta_j$. For $t_i \in T$ and $m_j \in M$, the probability of having the correct response $P(y_{i,j} = 1)$ is defined by the item response function for the 2PL IRT model as:
\begin{equation}
\label{eq:irt-2pl}
\begin{aligned}
P(y_{i,j} = 1 \mid \theta_j, a_i, b_i) &= \sigma(a_i(\theta_j - b_i)) \\
&= \frac{e^{a_i(\theta_j - b_i)}}{1 + e^{a_i(\theta_j - b_i)}}
\end{aligned}
\end{equation}
where $\sigma$ is the logistic function, and $a_i$, $b_i$, and $\theta_j$ are learned parameters.

Intuitively, task $t_i$ is not informative when $a_i = 0$. Both $b_i$ and $\theta_j$ are scale invariant: their absolute values are not identifiable, and only their difference $\theta_j - b_i$ matters for the probability of success. Hence $b_i$ and $\theta_j$ are centered at zero and are on the same scale. 
When $P(y_{i,j} = 1 \mid \theta_*, a_i, b_i)=0.5$, $\theta_* = b_i$, unless $a_i = 0$. $\theta_*$ naturally provides the definition for the ability required to achieve a 50\% success rate on a specific question with difficulty $b_i$.

IRT models have recently been adapted to language model evaluation to characterize task difficulty and model capability from performance data. \citet{Lalor2019LearningLP} introduced the use of IRT models from machine-generated responses rather than human responses to estimate the difficulty of NLP tasks. \citet{Rodriguez2021EvaluationEA} proposed Difficulty and Ability Discriminating (DAD) leaderboards, using IRT to infer task difficulty, discriminativeness, and feasibility. More recently, Fluid Benchmarking \citep{hofmann2025fluidlanguagemodelbenchmarking} applied IRT to estimate latent model ability and item difficulty and combined it with adaptive item selection to reduce variance and improve benchmarking efficiency and validity.

IRT maps performance to a shared latent space, enabling principled comparison and joint estimation of task difficulty and model capability. However, because IRT is inherently scale-invariant, its latent difficulty and ability parameters are identifiable only up to an affine transformation and therefore lack intrinsic semantic meaning. Consequently, although IRT offers a useful psychometric structure, it does not yield an interpretable measure of task difficulty in human terms or model capability relative to real-world effort.

We address this limitation by anchoring the IRT latent difficulty scale to human completion time, resolving the scale ambiguity and endowing IRT difficulty and ability estimates with a concrete, human-centric interpretation that enables prediction of human task duration from model performance.


\section{Predicting Human Task Time with BRIDGE}
\label{sec:method}

While direct human studies provide expensive but valuable estimates of task difficulty in human terms, IRT offers a scalable but intrinsically uncalibrated measure of task difficulty from model performance. We introduce BRIDGE, a framework that connects these two regimes by anchoring latent IRT difficulty to human completion time.

Given binary outcomes for each model–task pair, we fit a 2PL IRT model and estimate the discrimination $a_i$ and difficulty $b_i$ for each task, along with an ability $\theta_j$ for each model. The IRT model is fitted to maximize the likelihood $\prod_{i,j} P\!\left(y_{ij} \mid \theta_j, a_i, b_i\right)$ using Markov chain Monte Carlo with hierarchical priors, based on \citet{natesan2016bayesian}.\footnote{We use the implementation from \citet{Lalor_2023}.} The binary run matrix can be sparse due to limited compute and incomplete historical evaluations, but the observed responses (51\% in our experiments) are sufficient to support stable estimation under the 2PL model.

Expressing model capability in terms of the odds $P/(1-P)$, we can rewrite \cref{eq:irt-2pl} in log-odds form, 
\begin{equation}
\label{eq:log-odd}
\log \frac{P}{1 - P} = a_i(\theta_j - b_i)
\end{equation}
Thus, the latent ability parameter $\theta$ is a log-scale measure of model capability. If capability grows exponentially over time $t$ as prior scaling analyses suggest, then $\theta$ grows linearly with $t$. 
Independently, METR's results show that the human task-time horizon $h$ also grows exponentially with $t$ (i.e., $h \propto e^t$, equivalently $t \propto \log h$). Combining these two observations: $\theta \propto \log h$. 
Because task difficulty $b_i$ is defined on the same latent scale as $\theta$, this implies a linear relationship between task difficulty $b_i$ and the logarithm of the human time required to complete task $t_i$.

Let $T_h \subseteq T$ denote the subset of tasks with human completion time annotations. For each task $t_k \in T_h$, let $b_k$ denote its latent task difficulty and $h_k$ its human completion time. As the task difficulty is on the $\log$ scale of human completion time, we can fit a linear relationship between $\{ b_k : t_k \in T_h \}$ and $\{ \log{h_k} : t_k \in T_h \}$: 
\begin{equation}
\label{eq:humans}
\log(h_k) = \text{slope} \times b_k + \text{intercept}
\end{equation}
This mapping enables prediction of human completion time for tasks in $T_{h^\prime} \not\subset T$ using only their IRT latent difficulties, without requiring task descriptions or human studies.

Building on this calibration, we forecast the evolution of frontier model capability in human-interpretable units. Specifically, within each 2-month release window, we identify the best-performing model $m_s$ with its estimated ability $\theta_s$. At a 50\% success rate, $b_s = \theta_s$ characterizes the maximum task difficulty solvable. More generally, when the success rate is not 50\%, given model capability $\theta_s$, we find tasks with $a_s$ and $b_s$, which leads to the desired success probability with the 2PL model (\Cref{eq:irt-2pl}).\footnote{Primary analysis focuses on 50\% success threshold. A detailed example for the 80\% success rate is discussed in \Cref{subsec:forecast_p80}.} Applying the calibration in \Cref{eq:humans} on $b_s$ then yields the associated human completion time $h_s$, producing time-horizon forecasts of model capability that are directly anchored to human task duration.


\section{Experiments}
\label{sec:experiments}

We use METR to learn the calibration mapping between latent task difficulty $b$ and human completion time. Concretely, after fitting the 2PL IRT model on the response matrix from all benchmarks, we regress $\log h_i$ on $b_i$ to obtain the log-linear mapping used throughout the paper.

We apply this calibration to estimate human completion time on benchmarks without exact time annotations. We validate this on a set of out-of-distribution benchmarks, which require extended multi-step reasoning or workflow execution. We use SWE-Bench verified \cite{jimenez2024swebench}, MLE-bench \cite{chan2025mlebenchevaluatingmachinelearning}, GDPval \cite{patwardhan2025gdpvalevaluatingaimodel}, and Cybench \cite{zhang2025cybench}, to test how well we can estimate human completion time and to forecast trends without human time annotations.

\subsection{Data}

\subsubsection{METR (calibration benchmark)}
\label{subsubsec:datametr}

METR \citep{kwa2025measuringaiabilitycomplete} released a unified dataset aggregating model performance on 170 tasks spanning three benchmarks: Software Atomic Actions (SWAA), which comprise single-step computer operation tasks; HCAST, which focuses on software engineering tasks \citep{rein2025hcasthumancalibratedautonomysoftware}; and RE-Bench, which evaluates AI research and engineering tasks \citep{wijk2025rebenchevaluatingfrontierai}. Each task is annotated with \emph{how long a domain-knowledgeable human without task-specific context} would take to complete it.

METR reports repeated trials for each model--task pair. To obtain a single binary outcome per pair compatible with our IRT fit, we treat a model-task pair as successful if the model succeeds in at least 50\% of the reported trials.

\subsubsection{Out-of-Distribution Validation Benchmarks}
\label{subsubsec:oodbenchmarks}

\paragraph{SWE-bench Verified:} 
SWE-bench \citep{jimenez2024swebench} evaluates software engineering capability by measuring whether an agent can resolve real GitHub issues. SWE-bench Verified is a curated subset of 500 instances that have been manually validated and annotated with coarse completion-time categories: \textless 15 minutes, 15 minutes to 1 hour, 1 hour to 4 hours, and \textgreater 4 hours. 
A task is successful if the generated patch resolves the corresponding GitHub issue, as verified by passing the benchmark evaluation harness (i.e., the relevant unit tests).
In addition to our own runs, we incorporate publicly available SWE-bench Verified leaderboard logs, which provide task-level success outcomes for multiple model--agent configurations. 

\paragraph{MLE-bench:}
MLE-bench \citep{chan2025mlebenchevaluatingmachinelearning} evaluates the capabilities of LLMs on end-to-end machine learning workflows, including data preparation, model training, and hyperparameter tuning. The benchmark comprises 75 Kaggle competitions spanning diverse domains and problem types, with tasks categorized into difficulty levels ranging from low to hard. 
For each competition, we construct three binary success indicators: (i) whether the agent produces a valid submission under the competition rules, (ii) whether the submission achieves above-median performance on the public leaderboard, and (iii) whether it earns any Kaggle medal (bronze/silver/gold). We treat these indicators as three separate tasks per competition and rely on the official evaluation pipelines.
We select a subset of 38 problems, consisting of 20 low difficulty tasks and 18 medium or hard difficulty tasks, due to computational constraints.

\paragraph{GDPval:}
GDPval \citep{patwardhan2025gdpvalevaluatingaimodel} evaluates economically significant, real-world tasks spanning 44 occupations across nine major U.S. GDP-contributing sectors. Tasks are designed by industry professionals (average 14 years of experience) to reflect real workflows, following occupational definitions from the U.S. Bureau of Labor Statistics. 
Because GDPval is open-ended, it does not admit a fully automated, verifiable evaluator. We therefore use an LLM-as-a-judge pipeline with Gemini 3 Pro as the judge. 

\paragraph{Cybench:}
Cybench \citep{zhang2025cybench} evaluates cybersecurity capabilities by measuring whether agents can solve professional-level Capture the Flag (CTF) challenges. The benchmark comprises tasks drawn from four CTF competitions spanning cryptography, forensics, reverse engineering, web exploitation, and binary exploitation, with difficulty levels ranging from very easy to hard, and human first-solve times (FST) from under 5 minutes to over 24 hours. A task is deemed successful if the agent recovers the correct flag string, verified by exact match. We run agents using the Cybench agent framework\footnote{\url{https://github.com/andyzorigin/cybench}} with tool access to a bash shell and standard CTF utilities. We use the unguided setting and restrict analysis to the tasks for which at least one model succeeds, as tasks unsolved by all models yield unreliable IRT difficulty estimates and would not support meaningful human time prediction.

For SWE-bench Verified, MLE-bench, and GDPval, we run agents using the InspectAI framework\footnote{\url{https://inspect.aisi.org.uk/}} with the default ReAct-style scaffold and tool access to a bash shell and python interpreter. For each task, generation is capped at 1000 turns or until context window is full. We use a temperature of 0.0 (greedy decoding) and sufficiently large maximum output token budgets to avoid truncation of intermediate reasoning steps or final outputs.

The included models are described in \Cref{app:models-list} and additional evaluation details are described in \Cref{app:eval_methods}.

\subsection{Baselines}

\subsubsection{Logit Success Rate}
\label{sec:baseline}

Beside the probabilistic IRT model, we include a heuristic baseline that estimates task difficulty and model ability via success rates. Such accuracy-based methods are widely used and offer an intuitive notion of task difficulty and model capability. Let $Y \in \{0,1\}^{|T| \times |M|}$ denote the binary response matrix of observed model–task outcomes, with rows corresponding to tasks and columns to models.

For each task $t_i$, we compute its empirical success rate across models over observed entries,
$r_i = \frac{1}{|M_i|} \sum_{j \in M_i} Y_{ij}$,
where $M_i$ denotes the set of models with available responses for task $i$. We define the baseline difficulty score as
$d_i^{\text{base}} = -\operatorname{logit}(1 - r_i)$,
where $\operatorname{logit}(p) = \log \frac{p}{1-p}$.

Analogously, for each model $m_j$, we compute its empirical success rate across tasks, $s_j = \frac{1}{|T|} \sum_i Y_{ij}$, and define its baseline ability as
$a_j^{\text{base}} = \operatorname{logit}(s_j)$.

We apply the logit transform to both task- and model-level average success rates to place the scores on a unbounded scale that is comparable to the log-odds structure of the 2PL IRT model. To enable human task completion time prediction, we fit the same log-linear calibration as in \Cref{eq:humans} by regressing $\log h_i$ on $d_i^{\text{base}}$ for METR tasks with human time annotations. This yields a mapping from success rate based difficulty to human completion time.

We use this baseline as a point of comparison for BRIDGE in task-time estimation to assess the added value of explicitly modeling discrimination and latent ability via IRT.

\subsubsection{LLMs as Estimators}

Prior works has shown that LLMs can be used to estimate both the latent difficulty \citep{razavi2025estimatingitemdifficultyusing, tabib2025trustworthydifficultyassessmentslarge} and human task completion time \citep{veeramani-etal-2024-large}.
As an additional baseline, we prompt frontier LLMs (Gemini 3 Pro and GPT-5.2) to predict human completion time for tasks from task descriptions.

For each task, we construct a prompt that provides task-specific context and requests a point estimate of expert human completion time in minutes. While adapted to the domain characteristics of each benchmark, all prompts follow a consistent schema: (1) relevant metadata (e.g., repository or competition information), (2) the full task description or problem statement, (3) domain-appropriate calibration anchors spanning the expected difficulty range, and (4) explicit instructions to output a single numeric estimate accompanied by a structured justification.
We use greedy decoding in a single turn and a maximum output token budget of 32000. 
The full prompt templates are shown in \Cref{fig:swebench-time-estimation-prompt,fig:cybench-time-estimation-prompt}.

\subsection{Estimating Human Task Completion Time}

\begin{figure}
    \centering
    \includegraphics[width=0.6\linewidth]{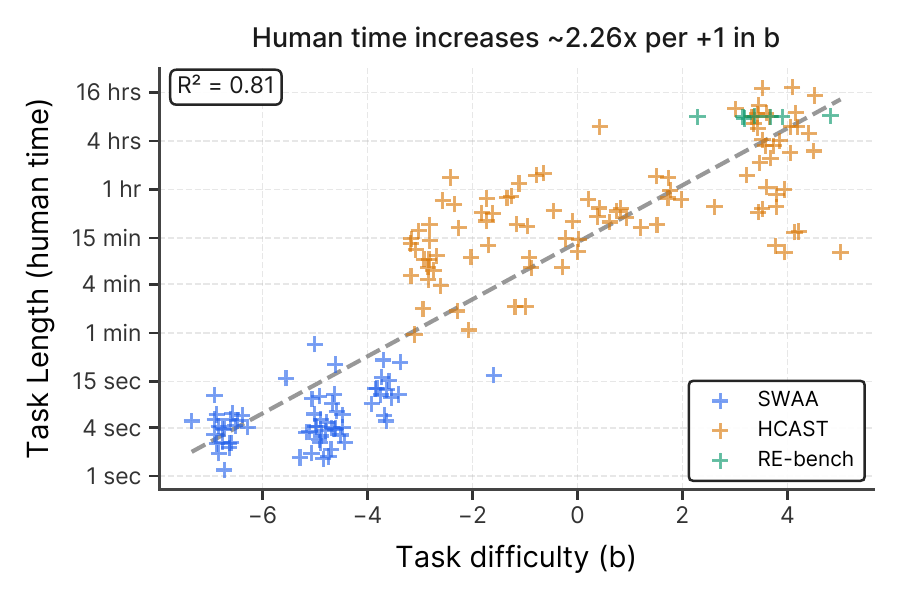}
    \caption{Task length (human completion time) vs. latent task difficulty ($b$) estimated via 2PL IRT across METR task suites (SWAA, HCAST, RE-bench), based on \Cref{eq:humans}. The log-linear fit ($R^2 = 0.81$) shows that each unit increase in $b$ corresponds to $\sim2.26\times$ increase in human completion time. This calibration anchors the IRT latent difficulty scale to human-interpretable units to enable prediction of task duration directly from model performance.}
    \label{fig:time_vs_b}
\end{figure}

After fitting the 2PL IRT model on the response matrix from all benchmarks, using the METR aggregated dataset, we fit a linear model relating latent task difficulty $b_i$ to the logarithm of human task completion time $\log h_i$ (in minutes), as defined in \Cref{eq:humans} and shown in \Cref{fig:time_vs_b}. With $R^2 = 0.81$, the fitted model indicates a strong log-linear alignment between latent difficulty and human completion time. According to the fitted slope, a one-unit increase in difficulty $b$ corresponds to about a $2.26\times$ increase in human completion time.

Having established the log-linear relationship between latent task difficulty $b$ and human completion time $h$, we apply this mapping to estimate human task durations for tasks without exact time annotations: SWE-bench Verified, MLE-bench, GDPval, and Cybench. For each task, we infer task length $h_i$ from its IRT latent difficulty $b_i$ using the learned log-linear relationship. \Cref{fig:task_length_estimation} in \Cref{app:task-length-estimation} shows the task-length distributions.

\begin{figure*}[t]
    \centering
    \includegraphics[width=\linewidth]{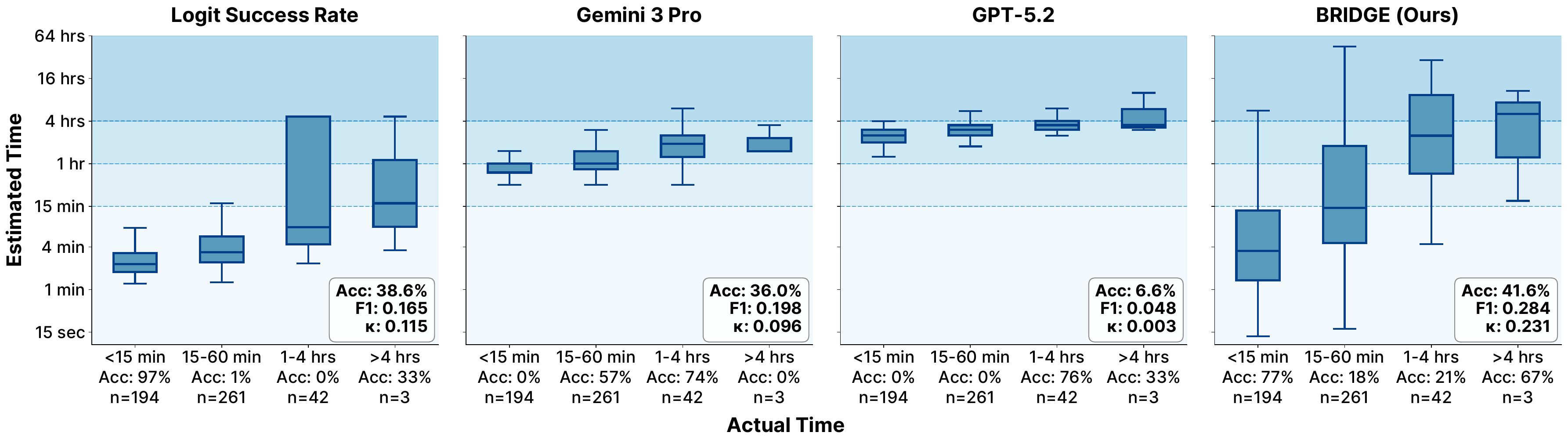}
    \caption{Alignment between annotated human completion time buckets and estimated human completion times on SWE-bench Verified. We report per-bucket classification accuracy (Acc) and the number of tasks (n), as well as overall accuracy, weighted macro F1 score, and weighted kappa. We compare a logit success-rate heuristic, LLM-based time predictions (Gemini 3 Pro, GPT-5.2), and BRIDGE. BRIDGE achieves substantially better alignment with the annotated time buckets than both heuristic and LLM-based baselines.}
    \label{fig:swebench_forecast}
\end{figure*}

\begin{figure*}[t]
    \centering
    \includegraphics[width=\linewidth]{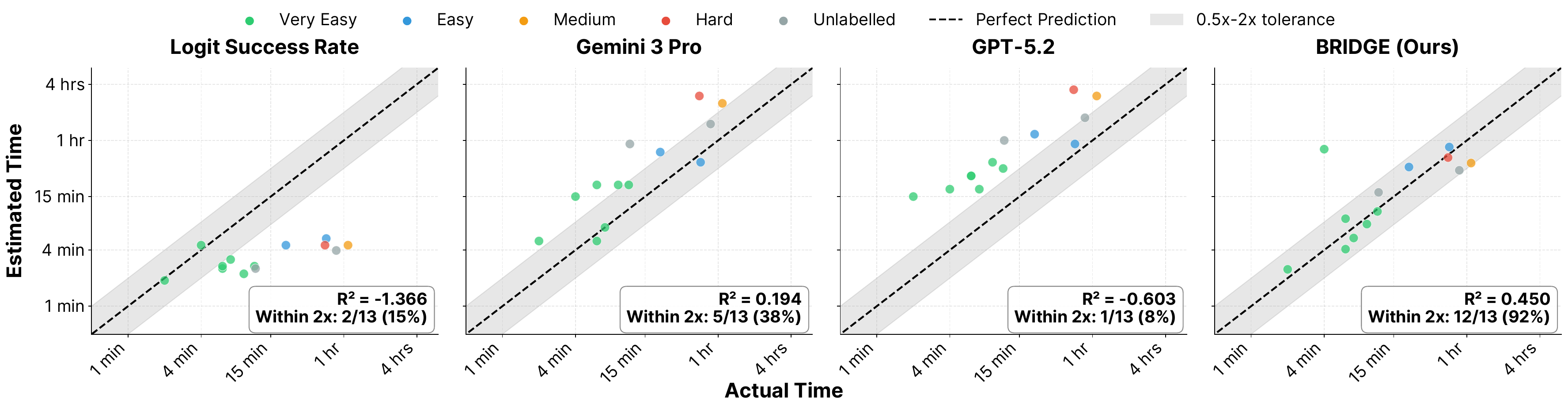}
    \caption{Alignment between actual human completion time (first-solve time) and estimated completion times on Cybench. The logit success-rate baseline substantially underestimates task duration, while LLM-based estimates consistently overestimate it. In contrast, BRIDGE aligns closely with actual human times, with 92.3\% of tasks falling within a $0.5\times$--$2\times$ tolerance band.}
    \label{fig:cybench_forecast}
\end{figure*}

We evaluate the transferability of this calibration on SWE-bench Verified and Cybench by comparing predicted times against human-annotated completion-time. On SWE-bench Verified (\Cref{fig:swebench_forecast}), BRIDGE produces predictions that increase monotonically across annotated buckets, with median estimates that track the bucket ordering and scale. Despite unavoidable discretization error introduced by coarse bucket boundaries, BRIDGE achieves substantially better alignment with the annotated time buckets than both heuristic and LLM-based baselines.
In contrast, the logit success-rate baseline systematically underestimates task duration, while Gemini 3 Pro and GPT-5.2 yield compressed predictions with limited dynamic range, leading to poor separation between longer-horizon buckets.

On Cybench (\Cref{fig:cybench_forecast}), BRIDGE’s predicted human completion time estimates align closely with recorded human first-solve times, achieving the strongest correlation ($R^2=0.45$) and placing 92.3\% of tasks within a $0.5\times - \; 2\times$ tolerance band. The logit success rate baseline substantially underestimates task duration, reflecting its inability to account for heterogeneity in model capability. Gemini 3 Pro and GPT-5.2 capture the correct qualitative ordering of tasks but consistently overestimate absolute completion times, resulting in markedly lower tolerance-band coverage.

On MLE-bench (\Cref{fig:task_length_estimation}), tasks yielding only valid submissions are estimated to require substantially shorter completion times than achieving above median performance on the leaderboard or earning any medal.

Overall, these results show that BRIDGE yields human task completion time estimates that are quantitatively accurate where ground truth is available and qualitatively well calibrated across benchmarks, consistently outperforming success rate based and LLM based human completion time prediction baselines.

\subsection{Forecasting Capabilities Without Human Studies}

\begin{figure*}[t]
    \centering
    \includegraphics[width=\linewidth]{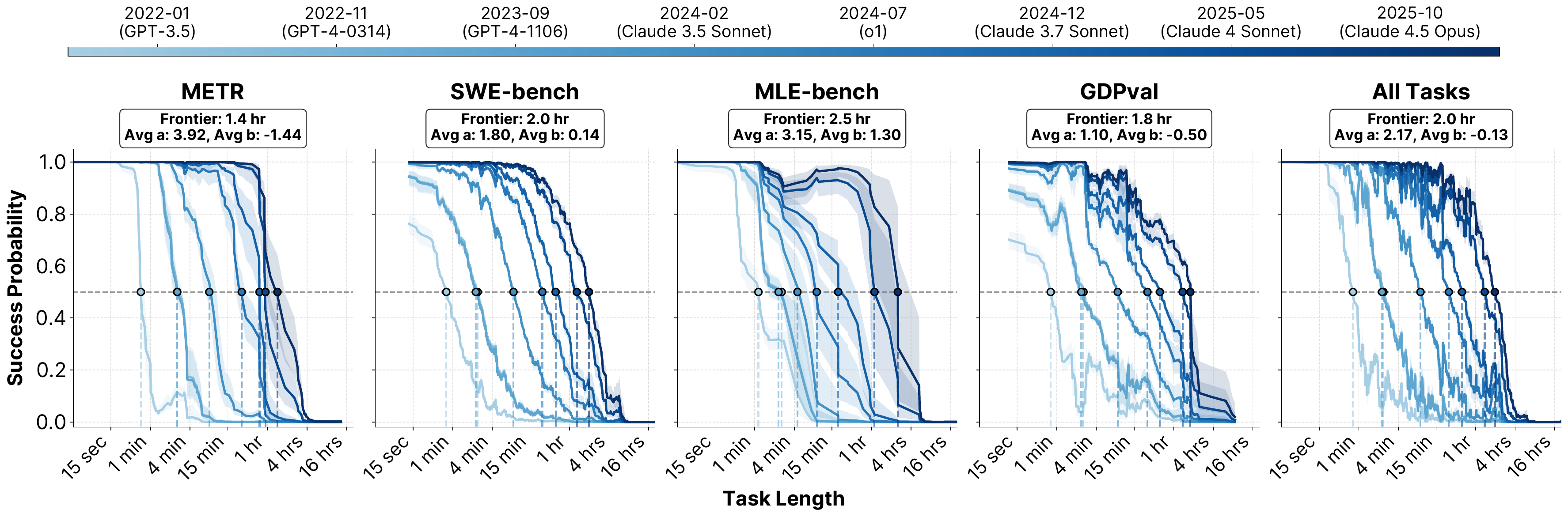}
    \caption{Success probability versus estimated human task completion time for different models, smoothed with a window of 15 tasks. Solvable task lengths at the 50\% success threshold are indicated across model release dates, with darker blue denoting more recent models. SOTA models achieve 50\% success on tasks estimated to require $\sim$1.4–2.5 hours of human effort. Steeper curves reflect higher task discrimination parameters $a$. Non-smoothness arises from heterogeneity in task-level difficulty and discrimination $(a_i, b_i)$, highlighting the importance of task-level granularity.}
    \label{fig:inverse-sigmoid-p50}
\end{figure*}

To track the evolution of frontier model capabilities, we analyze task success probability as a function of estimated human completion time on a logarithmic scale. We partition models into 5-month release windows and, within each window, select the model with the highest estimated ability, denoted $m_{\max}$ with ability $\theta_{\max}$. For tasks in METR, SWE-bench Verified, MLE-bench, and GDPval, we use the IRT-estimated latent parameters $(a_i, b_i)$ together with $\theta_{\max}$ to compute task success probabilities $p_i$ via \Cref{eq:irt-2pl}. Each task’s latent difficulty $b_i$ is then mapped to estimated human completion time $h_i$ using \Cref{eq:humans}. We plot $p_i$ against $h_i$ across tasks in \Cref{fig:inverse-sigmoid-p50}.

Frontier models released in 2025 achieve 50\% success on tasks estimated to require $\sim 1$--$2.5$ hours of human effort. Raising the success threshold from 50\% to 80\% substantially contracts the range of solvable task lengths: as shown in \Cref{fig:inverse-sigmoid-p80}, even the most recent models are largely limited to tasks requiring less than one hour of human completion time. The plotted probability–time relationships are not perfectly smooth inverse sigmoids due to heterogeneity in task discrimination $a_i$ and difficulty $b_i$ even within the same benchmark, underscoring the importance of task-level granularity. We further observe markedly steeper curves for METR and SWE-bench than for MLE-bench and GDPval, consistent with their higher discrimination parameters $a$.

\begin{figure*}[t]
    \centering
    \begin{subfigure}[t]{0.48\linewidth}
        \centering
        \includegraphics[width=\linewidth]{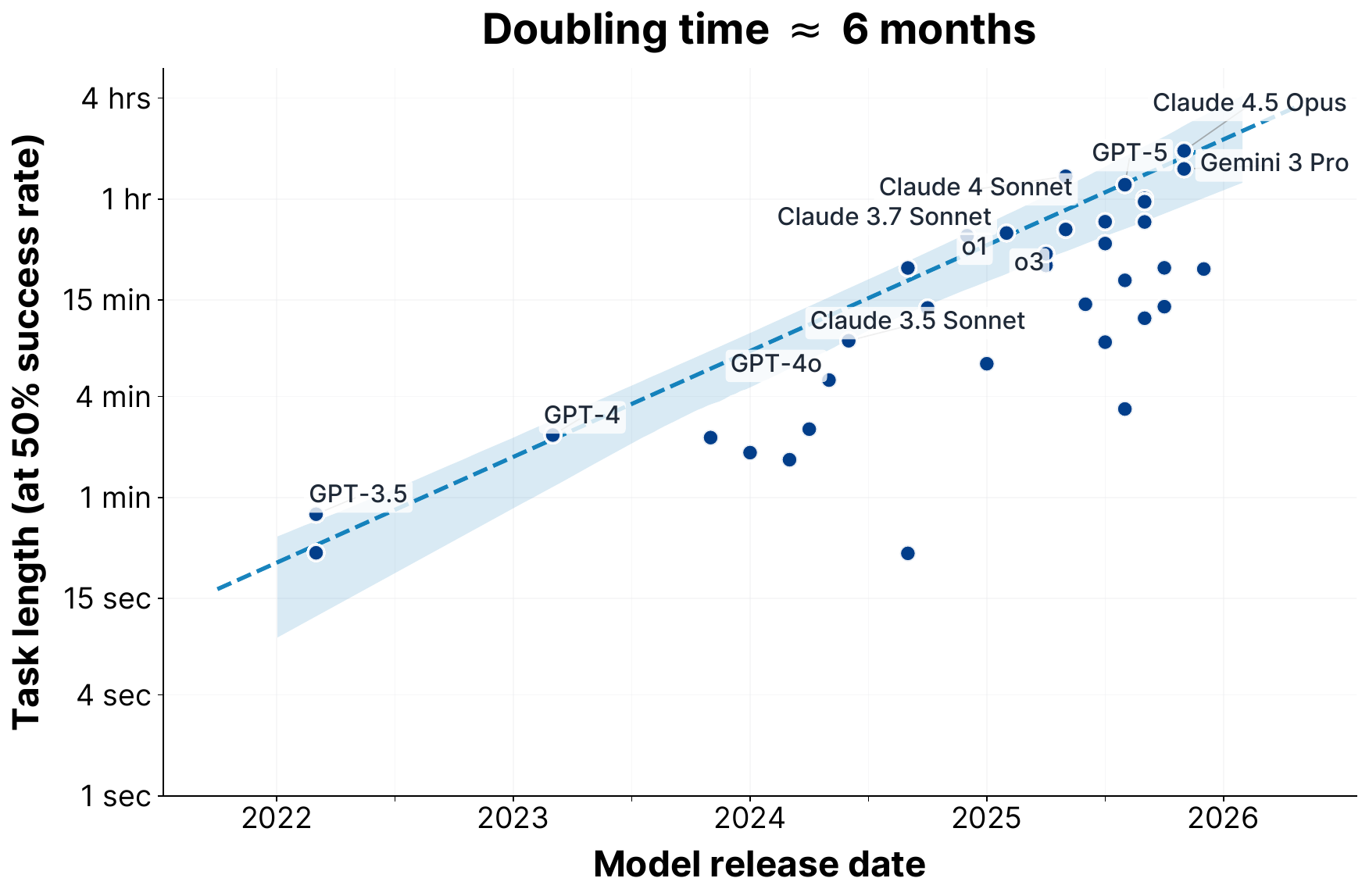}
    \end{subfigure}\hfill
    \begin{subfigure}[t]{0.48\linewidth}
        \centering
        \includegraphics[width=\linewidth]{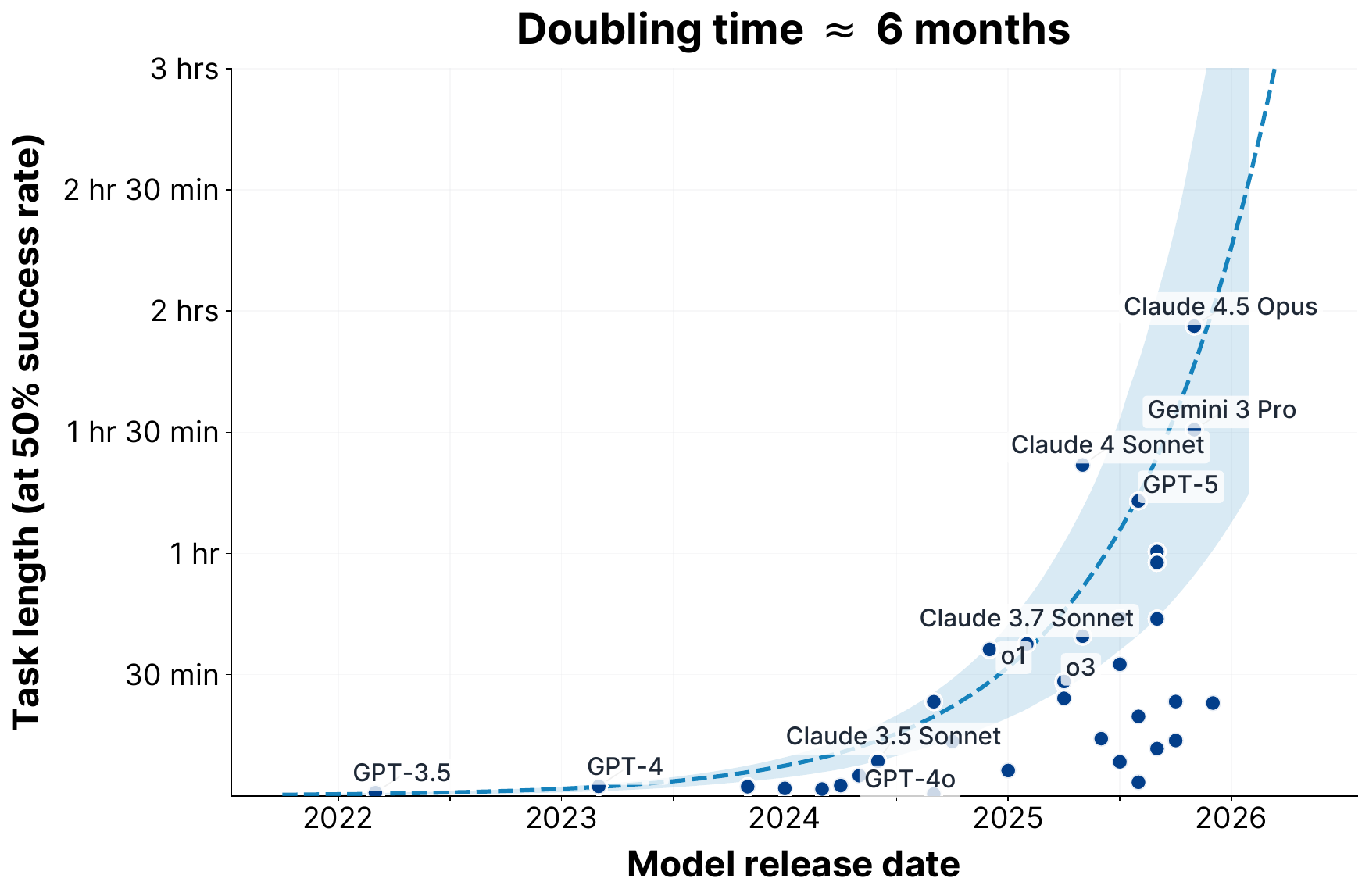}
    \end{subfigure}
    \caption{Forecasting trends of task length horizon over model release date without human task-time annotation. The task length at which model can achieve 50\% accuracy grows exponentially over time, with an estimated doubling time of approximately 6 months. The left subfigure shows this trend on a logarithmic scale for task length while the right subfigure presents the same trend on a linear scale. The shaded region represents 95\% confidence intervals computed via bootstrapping. BRIDGE enables forecasting of frontier model capabilities in human-interpretable units using model performance alone.}
    \label{fig:forecast_p50}
\end{figure*}

Finally, we use BRIDGE to forecast how the human completion time horizon of frontier models evolves over model release dates without human annotations. We partition models into 2-month release windows and select the best-performing model $m_s$ in each window with estimated ability $\theta_s$. For 50\% success rate, $b_s = \theta_s$. 
$b_s$ is then mapped to human completion time $h_s$ with \Cref{eq:humans}. \Cref{fig:forecast_p50} visualizes task length horizon $h_i$ vs. model release date.
At a 50\% success rate, frontier models exhibit exponential growth in solvable task length over time, with capabilities doubling approximately every $6$ months. 
The current state-of-the-art (SOTA) model reaches a task-length horizon of roughly two hours at 50\% success rate. 
We observe analogous behavior at an 80\% success threshold (\Cref{fig:forecast_p80}), albeit with a reduced achievable task length and a similar doubling time.

Overall, these results show that BRIDGE enables forecasting of frontier model capabilities in human-interpretable units using model performance alone. The resulting growth rates closely align with estimates obtained from direct human studies, as in \citet{kwa2025measuringaiabilitycomplete}, supporting the use of model performance-derived latent difficulty as a scalable and reliable proxy for human time annotations when tracking long-horizon capability progress.

\subsection{Sensitivity to model pool and sparsity}

As models are not evaluated on all tasks, we assess robustness to incomplete model-task coverage via a sparsity ablation, randomly removing 10\% to 70\% of observed entries and refitting IRT. Sparsity (\%) denotes the fraction of originally observed model-task responses removed, and Observed (\%) denotes the fraction of the full model-task matrix that remains observed after ablation. $\rho(b)$ is the Spearman correlation between ablated-data task difficulty estimates and the full-data estimates, and $\rho(\theta)$ is the corresponding correlation for model ability.

\begin{table}[h]
\centering
\small
\caption{Robustness of IRT parameter estimates under different sparsity levels in model-task binary success indicator response matrix. The actual estimates use 51\% observed values in the response matrix.}
\begin{tabular}{cccc}
\toprule
\textbf{Sparsity (\%)} & \textbf{Observed (\%)} & $\boldsymbol{\rho(b)}$ & $\boldsymbol{\rho(\theta)}$ \\
\midrule
10 & 45 & 0.9947 & 0.9980 \\
20 & 40 & 0.9913 & 0.9970 \\
30 & 35 & 0.9869 & 0.9955 \\
40 & 30 & 0.9810 & 0.9947 \\
50 & 25 & 0.9735 & 0.9903 \\
60 & 20 & 0.9614 & 0.9872 \\
70 & 15 & 0.9424 & 0.9798 \\
\bottomrule
\end{tabular}
\label{tab:sparsity_ablation}
\end{table}

The latent difficulty parameter $b_i$ used for time calibration remains highly stable. Its correlation with the full-data estimate stays above 0.97 after removing 50\% of observations and above 0.94 even at 70\% removal. The latent model ability $\theta_j$ is even more stable, with consistently higher correlations across all sparsity levels. This suggests that the current 51\% observed values in the model-task response matrix is sufficient for reliable estimation, and BRIDGE is robust to different model-task coverages.


\section{Related Work}

\subsection{Estimating latent task difficulties}

Although fitting IRT models using model outputs requires significantly less effort than human annotators, it still requires a large and diverse set of tasks with a sufficient number of attempts. Consequently, a parallel line of research has focused on estimating latent task difficulty more efficiently, with the goal of reducing the number of required attempts. \citet{byrd-srivastava-2022-predicting} used traits correlated with features of questions, answers, and associated contexts, to predict both difficulty and discrimination for new questions. \citet{scarlatos-etal-2025-smart} estimate task difficulty by simulating responses from a population of LLM-based students, each conditioned on a scalar IRT ability parameter so that the model generates answers as if written by learners at different skill levels, automatically scoring these responses, and fitting an IRT model to infer item difficulty. \citet{truong2025reliableefficientamortizedmodelbased} trained a model that predicts question difficulty from its embedding features. This line of work is focusing on estimating the latent difficulty of newly added tasks and can be used in parallel with our work.

\subsection{Aligning human and latent task difficulty}

TaskSense \citep{yin2025tasksensecognitivechainmodeling} estimates task difficulty in the context of graphical user interface tasks by modeling the cognitive processes that precede each user action. It decomposes tasks into sequences of cognitive steps, 
and assigns each step a difficulty index based on information-theoretic and cognitive principles. Task difficulty is computed as a linear aggregation of these step-level difficulties. The resulting difficulty estimates are validated against human data and shown to correlate strongly with both step-level and task-level human completion times. Such an approach requires task-specific modeling and annotations. Our approach differs by treating task difficulty as a latent variable inferred from large-scale model performance across benchmarks, rather than from explicit cognitive decomposition, providing a unified generalizable difficulty estimation framework aligned with human completion time.
\subsection{Aligning human and latent model capability}

Concurrently, Epoch AI proposed an IRT-based framework to estimate benchmark difficulty and model capability \citep{ho2025rosettastoneaibenchmarks}. Unlike our Bayesian inference, they learned latent parameters with L2-regularized non-linear least-squares and derive the latent model ability score. They show that the latent model ability score correlates with METR human time across models and yields capability growth trends broadly consistent with ours.

Our approach differs in granularity and calibration: whereas Epoch AI operates at the benchmark level and anchor at the model level, we estimate difficulty at the individual task level and anchor it to human completion time, enabling task-level time prediction and horizon estimates. Our framework yields different predictions for certain frontier models compared to Epoch AI. For instance, Epoch AI ranks Gemini 3 Pro as the top-performing model, whereas our estimates place Claude 4.5 Opus ahead. We hypothesize that such differences likely stem from different benchmark compositions, highlighting the inherent uncertainty in capability forecasting. We believe that it is valuable to have multiple independent forecasting efforts: comparing predictions across methodologies can reveal where estimates are robust and where greater caution is warranted, ultimately aimed to provide a more calibrated view of frontier model progress.


\section{Conclusion \& Future Work}

We present BRIDGE, a psychometric framework that aligns latent task difficulty inferred from model performance with human task completion time, enabling interpretable, scalable evaluation and forecasting of AI capabilities \textbf{without requiring human time annotations}. By fitting a 2PL IRT model to performance data across benchmarks, BRIDGE induces a unified latent scale that jointly captures task difficulty and model capability.

We show that latent task difficulty is strongly and linearly related to the logarithm of human completion time, allowing human task durations to be inferred for benchmarks lacking explicit timing annotations. Leveraging this alignment, we characterize model success probability as a function of task length and find that recent frontier models achieve 50\% success on tasks requiring approximately $2$ hours of human effort. Tracking frontier models over time, BRIDGE reveals an exponential expansion of the solvable human task-length horizon, with capabilities doubling every 6 months, reproducing and corroborating METR’s human-time–based findings \textbf{using only model performance data}.

Looking forward, BRIDGE provides a foundation for a broader class of human-interpretable evaluations. One promising direction is to extend the framework to human–AI collaborative settings, modeling how task responsibility is dynamically shared, how partial automation reshapes effective task difficulty, and how human interventions can act as signals of difficulty as model capabilities advance.

BRIDGE can also be expanded beyond long-horizon procedural tasks to support knowledge-intensive evaluations by enriching the psychometric model with task attributes that capture information requirements, external tool use, or reliance on knowledge beyond a model’s training cutoff.

Finally, as task horizons lengthen, human completion times are expected to exhibit increased inter-individual variability. Accounting for this variability offers a natural direction for extending BRIDGE toward uncertainty-aware difficulty estimation and capability forecasting.

Overall, BRIDGE demonstrates that robust, human-grounded, and interpretable capability evaluation and forecasting can be achieved \textbf{without direct human studies}, substantially lowering the cost and friction of tracking real-world AI progress while remaining consistent with human-time–anchored evaluations.

\newpage

\section*{Acknowledgements}

Siva Reddy and Dzmitry Bahdanau are supported by the Canada CIFAR AI Chairs program. We acknowledge the support of the IVADO R3AI Grant and a Gemini Academic Program Award. We thank the Mila IDT team and the Digital Research Alliance of Canada for providing the computational resources used in this work. We also thank members of McGill University and Mila, especially Arkil Patel and Parishad BehnamGhader, for their valuable feedback and constructive discussions throughout the project.



\bibliography{main}

@book{baker2001basics,
  title={The basics of item response theory},
  author={Baker, Frank B},
  year={2001},
  publisher={ERIC}
}

@inproceedings{hofmann2025fluidlanguagemodelbenchmarking,
    title={Fluid Language Model Benchmarking},
    author={Valentin Hofmann and David Heineman and Ian Magnusson and Kyle Lo and Jesse Dodge and Maarten Sap and Pang Wei Koh and Chun Wang and Hannaneh Hajishirzi and Noah A. Smith},
    booktitle={Second Conference on Language Modeling},
    year={2025}
}

@inproceedings{
    kwa2025measuringaiabilitycomplete,
    title={Measuring {AI} Ability to Complete Long Software Tasks},
    author={Thomas Kwa and Ben West and Joel Becker and Amy Deng and Katharyn Garcia and Max Hasin and Sami Jawhar and Megan Kinniment and Nate Rush and Sydney Von Arx and Ryan Bloom and Thomas Broadley and Haoxing Du and Brian Goodrich and Nikola Jurkovic and Luke Harold Miles and Seraphina Nix and Tao Roa Lin and Neev Parikh and David Rein and Lucas Jun Koba Sato and Hjalmar Wijk and Daniel M Ziegler and Elizabeth Barnes and Lawrence Chan},
    booktitle={The Thirty-ninth Annual Conference on Neural Information Processing Systems},
    year={2026},
    url={https://openreview.net/forum?id=CGNJL6CeV0}
}

@article{yin2025tasksensecognitivechainmodeling,
    title={TaskSense: Cognitive Chain Modeling and Difficulty Estimation for GUI Tasks},
    author={Yiwen Yin and Zhian Hu and Xiaoxi Xu and Chun Yu and Xintong Wu and Wenyu Fan and Yuanchun Shi},
    journal={arXiv preprint arXiv:2511.09309},
    year={2025},
    url={https://arxiv.org/abs/2511.09309},
}

@inproceedings{
    jimenez2024swebench,
    title={{SWE}-bench: Can Language Models Resolve Real-world Github Issues?},
    author={Carlos E Jimenez and John Yang and Alexander Wettig and Shunyu Yao and Kexin Pei and Ofir Press and Karthik R Narasimhan},
    booktitle={The Twelfth International Conference on Learning Representations},
    year={2024},
    url={https://openreview.net/forum?id=VTF8yNQM66}
}

@inproceedings{
    chan2025mlebenchevaluatingmachinelearning,
    title={{MLE}-bench: Evaluating Machine Learning Agents on Machine Learning Engineering},
    author={Jun Shern Chan and Neil Chowdhury and Oliver Jaffe and James Aung and Dane Sherburn and Evan Mays and Giulio Starace and Kevin Liu and Leon Maksin and Tejal Patwardhan and Aleksander Madry and Lilian Weng},
    booktitle={The Thirteenth International Conference on Learning Representations},
    year={2025},
    url={https://openreview.net/forum?id=6s5uXNWGIh}
}

@inproceedings{
    patwardhan2025gdpvalevaluatingaimodel,
    title={{GDP}val: Evaluating {AI} Model Performance on Real-World Economically Valuable Tasks},
    author={Tejal Patwardhan and Rachel Dias and Elizabeth Proehl and Grace Kim and Michele Wang and Olivia Watkins and Simon Posada Fishman and Marwan Aljubeh and Phoebe Thacker and Laurance Fauconnet and Natalie S. Kim and Samuel Miserendino and Gildas Chabot and David Li and Patrick Chao and Michael Sharman and Alexandra Barr and Amelia Glaese and Jerry Tworek},
    booktitle={The Fourteenth International Conference on Learning Representations},
    year={2026},
    url={https://openreview.net/forum?id=hcuEdq6eKD}
}

@misc{t-AGI,
    author = {Richard Ngo},
    title = {Clarifying and predicting AGI},
    url = {https://www.lesswrong.com/posts/BoA3agdkAzL6HQtQP/clarifying-and-predicting-agi},
    year={2023},
}

@article{rein2025hcasthumancalibratedautonomysoftware,
    title={HCAST: Human-Calibrated Autonomy Software Tasks},
    author={David Rein and Joel Becker and Amy Deng and Seraphina Nix and Chris Canal and Daniel O'Connel and Pip Arnott and Ryan Bloom and Thomas Broadley and Katharyn Garcia and Brian Goodrich and Max Hasin and Sami Jawhar and Megan Kinniment and Thomas Kwa and Aron Lajko and Nate Rush and Lucas Jun Koba Sato and Sydney Von Arx and Ben West and Lawrence Chan and Elizabeth Barnes},
    journal={arXiv preprint arXiv:2503.17354},
    year={2025},
    url={https://arxiv.org/abs/2503.17354},
}

@inproceedings{
    wijk2025rebenchevaluatingfrontierai,
    title={{RE}-Bench: Evaluating Frontier {AI} R\&D Capabilities of Language Model Agents against Human Experts},
    author={Hjalmar Wijk and Tao Roa Lin and Joel Becker and Sami Jawhar and Neev Parikh and Thomas Broadley and Lawrence Chan and Michael Chen and Joshua M Clymer and Jai Dhyani and Elena Ericheva and Katharyn Garcia and Brian Goodrich and Nikola Jurkovic and Megan Kinniment and Aron Lajko and Seraphina Nix and Lucas Jun Koba Sato and William Saunders and Maksym Taran and Ben West and Elizabeth Barnes},
    booktitle={Forty-second International Conference on Machine Learning},
    year={2025},
    url={https://openreview.net/forum?id=3rB0bVU6z6}
}

@inproceedings{byrd-srivastava-2022-predicting,
    title = "Predicting Difficulty and Discrimination of Natural Language Questions",
    author = "Byrd, Matthew  and
      Srivastava, Shashank",
    editor = "Muresan, Smaranda  and
      Nakov, Preslav  and
      Villavicencio, Aline",
    booktitle = "Proceedings of the 60th Annual Meeting of the Association for Computational Linguistics (Volume 2: Short Papers)",
    month = may,
    year = "2022",
    address = "Dublin, Ireland",
    publisher = "Association for Computational Linguistics",
    url = "https://aclanthology.org/2022.acl-short.15/",
    doi = "10.18653/v1/2022.acl-short.15",
    pages = "119--130"
}

@inproceedings{scarlatos-etal-2025-smart,
    title = "{SMART}: Simulated Students Aligned with Item Response Theory for Question Difficulty Prediction",
    author = "Scarlatos, Alexander  and
      Fernandez, Nigel  and
      Ormerod, Christopher  and
      Lottridge, Susan  and
      Lan, Andrew",
    editor = "Christodoulopoulos, Christos  and
      Chakraborty, Tanmoy  and
      Rose, Carolyn  and
      Peng, Violet",
    booktitle = "Proceedings of the 2025 Conference on Empirical Methods in Natural Language Processing",
    month = nov,
    year = "2025",
    address = "Suzhou, China",
    publisher = "Association for Computational Linguistics",
    url = "https://aclanthology.org/2025.emnlp-main.1274/",
    doi = "10.18653/v1/2025.emnlp-main.1274",
    pages = "25082--25105",
    ISBN = "979-8-89176-332-6"
}

@inproceedings{
    truong2025reliableefficientamortizedmodelbased,
    title={Reliable and Efficient Amortized Model-based Evaluation},
    author={Sang T. Truong and Yuheng Tu and Percy Liang and Bo Li and Sanmi Koyejo},
    booktitle={Forty-second International Conference on Machine Learning},
    year={2025},
    url={https://openreview.net/forum?id=HDbWrsgkB9}
}

@article{Lalor2019LearningLP,
  title={Learning Latent Parameters without Human Response Patterns: Item Response Theory with Artificial Crowds},
  author={John P. Lalor and Hao Wu and Hong Yu},
  journal={Proceedings of the Conference on Empirical Methods in Natural Language Processing. Conference on Empirical Methods in Natural Language Processing},
  year={2019},
  volume={2019},
  pages={
          4240-4250
        },
  url={https://api.semanticscholar.org/CorpusID:201698284}
}

@inproceedings{Rodriguez2021EvaluationEA,
  title={Evaluation Examples are not Equally Informative: How should that change NLP Leaderboards?},
  author={Pedro Rodriguez and Joe Barrow and Alexander Miserlis Hoyle and John P. Lalor and Robin Jia and Jordan Lee Boyd-Graber},
  booktitle={Annual Meeting of the Association for Computational Linguistics},
  year={2021},
  url={https://api.semanticscholar.org/CorpusID:235703772}
}

@article{Lalor_2023,
   title={\texttt{py-irt}: A Scalable Item Response Theory Library for Python},
   volume={35},
   ISSN={1526-5528},
   url={http://dx.doi.org/10.1287/ijoc.2022.1250},
   DOI={10.1287/ijoc.2022.1250},
   number={1},
   journal={INFORMS Journal on Computing},
   publisher={Institute for Operations Research and the Management Sciences (INFORMS)},
   author={Lalor, John Patrick and Rodriguez, Pedro},
   year={2023},
   month=jan, pages={5–13} }

@article{natesan2016bayesian,
  title={Bayesian prior choice in IRT estimation using MCMC and variational Bayes},
  author={Natesan, Prathiba and Nandakumar, Ratna and Minka, Tom and Rubright, Jonathan D},
  journal={Frontiers in psychology},
  volume={7},
  pages={1422},
  year={2016},
  publisher={Frontiers}
}

@misc{how-does-time-horizon-vary-across-domains,
  title = {How Does Time Horizon Vary Across Domains?},
  author = {METR},
  url = {https://metr.org/blog/2025-07-14-how-does-time-horizon-vary-across-domains/},
  year = {2025},
  month = {07},
}

@article{ho2025rosettastoneaibenchmarks,
      title={A Rosetta Stone for AI Benchmarks}, 
      author={Anson Ho and Jean-Stanislas Denain and David Atanasov and Samuel Albanie and Rohin Shah},
      year={2025},
      journal={arXiv preprint arXiv:2512.00193},
      url={https://arxiv.org/abs/2512.00193}, 
}

@inproceedings{
zhang2025cybench,
title
=
{Cybench: A Framework for Evaluating Cybersecurity Capabilities and Risks of Language Models},
author
=
{Andy K Zhang and Neil Perry and Riya Dulepet and Joey Ji and Celeste Menders and Justin W Lin and Eliot Jones and Gashon Hussein and Samantha Liu and Donovan Julian Jasper and Pura Peetathawatchai and Ari Glenn and Vikram Sivashankar and Daniel Zamoshchin and Leo Glikbarg and Derek Askaryar and Haoxiang Yang and Aolin Zhang and Rishi Alluri and Nathan Tran and Rinnara Sangpisit and Kenny O Oseleononmen and Dan Boneh and Daniel E. Ho and Percy Liang},
booktitle
=
{The Thirteenth International Conference on Learning Representations},
year
=
{2025},
url
=
{https://openreview.net/forum?id=tc90LV0yRL},
}

@inproceedings{veeramani-etal-2024-large,
    title = "Large Language Model-based Pipeline for Item Difficulty and Response Time Estimation for Educational Assessments",
    author = "Veeramani, Hariram  and
      Thapa, Surendrabikram  and
      Shankar, Natarajan Balaji  and
      Alwan, Abeer",
    editor = {Kochmar, Ekaterina  and
      Bexte, Marie  and
      Burstein, Jill  and
      Horbach, Andrea  and
      Laarmann-Quante, Ronja  and
      Tack, Ana{\"i}s  and
      Yaneva, Victoria  and
      Yuan, Zheng},
    booktitle = "Proceedings of the 19th Workshop on Innovative Use of NLP for Building Educational Applications (BEA 2024)",
    month = jun,
    year = "2024",
    address = "Mexico City, Mexico",
    publisher = "Association for Computational Linguistics",
    url = "https://aclanthology.org/2024.bea-1.49/",
    pages = "561--566"
}

@article{razavi2025estimatingitemdifficultyusing,
      title={Estimating Item Difficulty Using Large Language Models and Tree-Based Machine Learning Algorithms}, 
      author={Pooya Razavi and Sonya J. Powers},
      year={2025},
      journal={arXiv preprint arXiv:2504.08804},
      url={https://arxiv.org/abs/2504.08804}, 
}

@article{tabib2025trustworthydifficultyassessmentslarge,
      title={Toward Trustworthy Difficulty Assessments: Large Language Models as Judges in Programming and Synthetic Tasks}, 
      author={H. M. Shadman Tabib and Jaber Ahmed Deedar},
      year={2025},
      journal={arXiv preprint arXiv:2511.18597},
      url={https://arxiv.org/abs/2511.18597}, 
}
\bibliographystyle{main}


\newpage
\appendix
\onecolumn


\section{List of Models}
\label{app:models-list}

\Cref{tab:model_coverage} lists all models and their evaluation coverage across task sources. We use the binary outcome (success or failure) of each model on each task across all these task sources to fit a 2PL IRT model, estimating item-level difficulty $b_i$ and discrimination $a_i$ as well as model ability $\theta_m$, as described in \Cref{sec:method}.

\begin{longtable}{p{6.5cm}cccccc}
\caption{Model evaluation coverage across task sources. A $\checkmark$ in each benchmark column indicates that evaluation logs for that model were available and included when fitting the 2PL IRT model. The ``Ours'' column denotes models we evaluated ourselves on the corresponding benchmarks; all remaining evaluation logs are derived from publicly available leaderboards.}
\label{tab:model_coverage} \\
\toprule
\shortstack{\phantom{X}\\\textbf{Model}} & \shortstack{\phantom{X}\\\textbf{METR}} & \shortstack{\textbf{SWE-}\\\textbf{bench}} & \shortstack{\phantom{X}\\\textbf{Cybench}} &
\shortstack{\phantom{X}\\\textbf{GDPval}} & \shortstack{\textbf{MLE-}\\\textbf{bench}} & \shortstack{\phantom{X}\\\textbf{Ours}} \\
\midrule
\endfirsthead
\multicolumn{7}{c}{\tablename\ \thetable{} -- \textit{Continued from previous page}} \\
\toprule
\shortstack{\phantom{X}\\\textbf{Model}} & \shortstack{\phantom{X}\\\textbf{METR}} & \shortstack{\textbf{SWE-}\\\textbf{bench}} & \shortstack{\phantom{X}\\\textbf{Cybench}} &
\shortstack{\phantom{X}\\\textbf{GDPval}} & \shortstack{\textbf{MLE-}\\\textbf{bench}} & \shortstack{\phantom{X}\\\textbf{Ours}} \\
\midrule
\endhead
\midrule
\multicolumn{7}{r}{\textit{Continued on next page}} \\
\endfoot
\bottomrule
\endlastfoot

\multicolumn{7}{l}{\textit{\textbf{OpenAI Models}}} \\
\midrule
GPT-2 & $\checkmark$ & $\times$ & $\times$ & $\times$ & $\times$ & $\times$ \\
GPT-3 & $\checkmark$ & $\times$ & $\times$ & $\times$ & $\times$ & $\times$ \\
GPT-3.5 Turbo Instruct & $\checkmark$ & $\times$ & $\times$ & $\times$ & $\times$ & $\times$ \\
GPT-4 (0314) & $\checkmark$ & $\times$ & $\times$ & $\times$ & $\times$ & $\times$ \\
GPT-4 (0125) & $\checkmark$ & $\times$ & $\times$ & $\times$ & $\times$ & $\times$ \\
GPT-4 (1106) & $\checkmark$ & $\checkmark$ & $\times$ & $\times$ & $\times$ & $\times$ \\
GPT-4 Turbo & $\checkmark$ & $\times$ & $\times$ & $\times$ & $\times$ & $\times$ \\
GPT-4o (2024-05-13) & $\checkmark$ & $\checkmark$ & $\checkmark$ & $\times$ & $\times$ & $\times$ \\
GPT-4o (2024-08-06) & $\times$ & $\checkmark$ & $\times$ & $\times$ & $\times$ & $\times$ \\
GPT-4o Mini & $\times$ & $\times$ & $\checkmark$ & $\checkmark$ & $\checkmark$ & $\checkmark$ \\
GPT-5 (2025-08-07) & $\times$ & $\checkmark$ & $\checkmark$ & $\checkmark$ & $\times$ & $\checkmark$ \\
GPT-5 Mini & $\times$ & $\checkmark$ & $\checkmark$ & $\checkmark$ & $\checkmark$ & $\checkmark$ \\
GPT-5 Nano & $\times$ & $\checkmark$ & $\checkmark$ & $\checkmark$ & $\checkmark$ & $\checkmark$ \\
GPT-5.1 & $\times$ & $\times$ & $\checkmark$ & $\checkmark$ & $\checkmark$ & $\checkmark$ \\
GPT-OSS-20B & $\times$ & $\checkmark$ & $\times$ & $\checkmark$ & $\checkmark$ & $\checkmark$ \\
GPT-OSS-120B & $\times$ & $\checkmark$ & $\times$ & $\checkmark$ & $\checkmark$ & $\checkmark$ \\
o1 & $\checkmark$ & $\times$ & $\times$ & $\times$ & $\times$ & $\times$ \\
o1-Preview & $\checkmark$ & $\checkmark$ & $\times$ & $\times$ & $\times$ & $\times$ \\
o3 & $\times$ & $\checkmark$ & $\checkmark$ & $\checkmark$ & $\checkmark$ & $\checkmark$ \\
o3-Mini & $\times$ & $\checkmark$ & $\times$ & $\times$ & $\times$ & $\times$ \\
o4-Mini & $\times$ & $\checkmark$ & $\checkmark$ & $\times$ & $\times$ & $\checkmark$ \\
\midrule

\multicolumn{7}{l}{\textit{\textbf{Anthropic Models}}} \\
\midrule
Claude 3 Opus & $\checkmark$ & $\checkmark$ & $\times$ & $\times$ & $\times$ & $\times$ \\
Claude 3.5 Sonnet (2024-06-20) & $\checkmark$ & $\times$ & $\times$ & $\times$ & $\times$ & $\times$ \\
Claude 3.5 Sonnet (2024-10-22) & $\checkmark$ & $\checkmark$ & $\times$ & $\times$ & $\times$ & $\times$ \\
Claude 3.7 Sonnet (2025-02-19) & $\checkmark$ & $\checkmark$ & $\times$ & $\times$ & $\times$ & $\times$ \\
Claude 4 Opus (2025-05-14) & $\times$ & $\checkmark$ & $\times$ & $\times$ & $\times$ & $\times$ \\
Claude 4 Sonnet (2025-05-14) & $\times$ & $\checkmark$ & $\times$ & $\times$ & $\times$ & $\times$ \\
Claude 4.5 Opus Medium & $\times$ & $\checkmark$ & $\times$ & $\times$ & $\times$ & $\times$ \\
Claude 4.5 Sonnet (2025-09-29) & $\times$ & $\checkmark$ & $\times$ & $\times$ & $\times$ & $\times$ \\
\midrule

\multicolumn{7}{l}{\textit{\textbf{Google Models}}} \\
\midrule
Gemini 2.5 Flash & $\times$ & $\times$ & $\times$ & $\checkmark$ & $\checkmark$ & $\checkmark$ \\
Gemini 2.5 Pro & $\times$ & $\checkmark$ & $\times$ & $\checkmark$ & $\checkmark$ & $\checkmark$ \\
Gemini 3 Pro Preview & $\times$ & $\checkmark$ & $\checkmark$ & $\checkmark$ & $\checkmark$ & $\checkmark$ \\
\midrule

\multicolumn{7}{l}{\textit{\textbf{Qwen Models}}} \\
\midrule
Qwen 2.5 Coder 32B & $\times$ & $\checkmark$ & $\times$ & $\checkmark$ & $\times$ & $\checkmark$ \\
Qwen 3 Coder 30B & $\times$ & $\checkmark$ & $\times$ & $\checkmark$ & $\checkmark$ & $\checkmark$ \\
Qwen 3 Coder 480B & $\times$ & $\checkmark$ & $\times$ & $\checkmark$ & $\checkmark$ & $\checkmark$ \\
\midrule

\multicolumn{7}{l}{\textit{\textbf{Other Foundation Models}}} \\
\midrule
DeepSeek V3.2 Reasoner & $\times$ & $\checkmark$ & $\times$ & $\times$ & $\times$ & $\times$ \\
GLM 4.5 & $\times$ & $\checkmark$ & $\times$ & $\times$ & $\times$ & $\times$ \\
GLM 4.6 & $\times$ & $\checkmark$ & $\times$ & $\times$ & $\times$ & $\times$ \\
Kimi K2 & $\times$ & $\checkmark$ & $\times$ & $\times$ & $\times$ & $\times$ \\
Llama 3.3 70B Instruct & $\times$ & $\times$ & $\times$ & $\checkmark$ & $\times$ & $\checkmark$ \\
MiniMax M2 & $\times$ & $\checkmark$ & $\times$ & $\checkmark$ & $\checkmark$ & $\checkmark$ \\
Entrpo Ekto 30B & $\times$ & $\checkmark$ & $\times$ & $\times$ & $\times$ & $\times$ \\
FrogBoss 32B & $\times$ & $\checkmark$ & $\times$ & $\times$ & $\times$ & $\times$ \\
FrogMini 14B & $\times$ & $\checkmark$ & $\times$ & $\times$ & $\times$ & $\times$ \\
\midrule

\multicolumn{7}{l}{\textit{\textbf{SWE-bench Agents}}} \\
\midrule
RAG + Claude 2 & $\times$ & $\checkmark$ & $\times$ & $\times$ & $\times$ & $\times$ \\
RAG + GPT-3.5 & $\times$ & $\checkmark$ & $\times$ & $\times$ & $\times$ & $\times$ \\
RAG + SWE-Llama 7B & $\times$ & $\checkmark$ & $\times$ & $\times$ & $\times$ & $\times$ \\
RAG + SWE-Llama 13B & $\times$ & $\checkmark$ & $\times$ & $\times$ & $\times$ & $\times$ \\
RAG + Claude 3 Opus & $\times$ & $\checkmark$ & $\times$ & $\times$ & $\times$ & $\times$ \\
RAG + GPT-4 & $\times$ & $\checkmark$ & $\times$ & $\times$ & $\times$ & $\times$ \\
Amazon Q Developer (Apr) & $\times$ & $\checkmark$ & $\times$ & $\times$ & $\times$ & $\times$ \\
MASAI + GPT-4o & $\times$ & $\checkmark$ & $\times$ & $\times$ & $\times$ & $\times$ \\
AppMap Navie + GPT-4o & $\times$ & $\checkmark$ & $\times$ & $\times$ & $\times$ & $\times$ \\
Factory Code Droid & $\times$ & $\checkmark$ & $\times$ & $\times$ & $\times$ & $\times$ \\
SWE-agent + Claude 3.5 Sonnet & $\times$ & $\checkmark$ & $\times$ & $\times$ & $\times$ & $\times$ \\
AutoCodeRover v2024-06-20 & $\times$ & $\checkmark$ & $\times$ & $\times$ & $\times$ & $\times$ \\
Amazon Q Developer (Jul) & $\times$ & $\checkmark$ & $\times$ & $\times$ & $\times$ & $\times$ \\
SWE-agent + GPT-4o & $\times$ & $\checkmark$ & $\times$ & $\times$ & $\times$ & $\times$ \\
EPAM AI Run + GPT-4o & $\times$ & $\checkmark$ & $\times$ & $\times$ & $\times$ & $\times$ \\
Honeycomb & $\times$ & $\checkmark$ & $\times$ & $\times$ & $\times$ & $\times$ \\
GRU (Aug) & $\times$ & $\checkmark$ & $\times$ & $\times$ & $\times$ & $\times$ \\
Lingma Agent 72B (Sep) & $\times$ & $\checkmark$ & $\times$ & $\times$ & $\times$ & $\times$ \\
Lingma Agent 7B (Sep) & $\times$ & $\checkmark$ & $\times$ & $\times$ & $\times$ & $\times$ \\
Solver (Sep 20) & $\times$ & $\checkmark$ & $\times$ & $\times$ & $\times$ & $\times$ \\
Solver (Sep 24) & $\times$ & $\checkmark$ & $\times$ & $\times$ & $\times$ & $\times$ \\
nFactorial (Oct 1) & $\times$ & $\checkmark$ & $\times$ & $\times$ & $\times$ & $\times$ \\
Lingma Agent 72B (Oct) & $\times$ & $\checkmark$ & $\times$ & $\times$ & $\times$ & $\times$ \\
Lingma Agent 7B (Oct) & $\times$ & $\checkmark$ & $\times$ & $\times$ & $\times$ & $\times$ \\
nFactorial (Oct 7) & $\times$ & $\checkmark$ & $\times$ & $\times$ & $\times$ & $\times$ \\
Composio SWEKit (Oct 16) & $\times$ & $\checkmark$ & $\times$ & $\times$ & $\times$ & $\times$ \\
Tools + Claude 3.5 Haiku & $\times$ & $\checkmark$ & $\times$ & $\times$ & $\times$ & $\times$ \\
Tools + Claude 3.5 Sonnet & $\times$ & $\checkmark$ & $\times$ & $\times$ & $\times$ & $\times$ \\
Emergent (Oct) & $\times$ & $\checkmark$ & $\times$ & $\times$ & $\times$ & $\times$ \\
Composio SWEKit (Oct 25) & $\times$ & $\checkmark$ & $\times$ & $\times$ & $\times$ & $\times$ \\
Solver (Oct 28) & $\times$ & $\checkmark$ & $\times$ & $\times$ & $\times$ & $\times$ \\
EPAM AI Run + Claude 3.5 Sonnet (Oct) & $\times$ & $\checkmark$ & $\times$ & $\times$ & $\times$ & $\times$ \\
nFactorial (Oct 30) & $\times$ & $\checkmark$ & $\times$ & $\times$ & $\times$ & $\times$ \\
nFactorial (Nov 5) & $\times$ & $\checkmark$ & $\times$ & $\times$ & $\times$ & $\times$ \\
Navie 2 + GPT-4o/Sonnet & $\times$ & $\checkmark$ & $\times$ & $\times$ & $\times$ & $\times$ \\
AutoCodeRover v2.0 + Claude 3.5 Sonnet & $\times$ & $\checkmark$ & $\times$ & $\times$ & $\times$ & $\times$ \\
Devlo (Nov) & $\times$ & $\checkmark$ & $\times$ & $\times$ & $\times$ & $\times$ \\
Nebius Search & $\times$ & $\checkmark$ & $\times$ & $\times$ & $\times$ & $\times$ \\
Artemis Agent & $\times$ & $\checkmark$ & $\times$ & $\times$ & $\times$ & $\times$ \\
Engine Labs & $\times$ & $\checkmark$ & $\times$ & $\times$ & $\times$ & $\times$ \\
MarsCode Agent & $\times$ & $\checkmark$ & $\times$ & $\times$ & $\times$ & $\times$ \\
SWE-Fixer + Qwen 2.5 72B (Nov) & $\times$ & $\checkmark$ & $\times$ & $\times$ & $\times$ & $\times$ \\
Agentless 1.5 + Claude 3.5 Sonnet & $\times$ & $\checkmark$ & $\times$ & $\times$ & $\times$ & $\times$ \\
Amazon Q Developer (Dec) & $\times$ & $\checkmark$ & $\times$ & $\times$ & $\times$ & $\times$ \\
GRU (Dec) & $\times$ & $\checkmark$ & $\times$ & $\times$ & $\times$ & $\times$ \\
EPAM AI Run + Claude 3.5 Sonnet (Dec) & $\times$ & $\checkmark$ & $\times$ & $\times$ & $\times$ & $\times$ \\
Google Jules + Gemini 2.0 Flash & $\times$ & $\checkmark$ & $\times$ & $\times$ & $\times$ & $\times$ \\
Devlo (Dec) & $\times$ & $\checkmark$ & $\times$ & $\times$ & $\times$ & $\times$ \\
CodeStory Midwit + Claude 3.5 Sonnet & $\times$ & $\checkmark$ & $\times$ & $\times$ & $\times$ & $\times$ \\
Emergent (Dec) & $\times$ & $\checkmark$ & $\times$ & $\times$ & $\times$ & $\times$ \\
BlackboxAI Agent v1.1 & $\times$ & $\checkmark$ & $\times$ & $\times$ & $\times$ & $\times$ \\
Learn by Interact + Claude 3.5 & $\times$ & $\checkmark$ & $\times$ & $\times$ & $\times$ & $\times$ \\
UGAIForge & $\times$ & $\checkmark$ & $\times$ & $\times$ & $\times$ & $\times$ \\
CodeShell Agent + Gemini 2.0 Flash & $\times$ & $\checkmark$ & $\times$ & $\times$ & $\times$ & $\times$ \\
Bracket & $\times$ & $\checkmark$ & $\times$ & $\times$ & $\times$ & $\times$ \\
AutoCodeRover v2.1 + Claude 3.5 Sonnet & $\times$ & $\checkmark$ & $\times$ & $\times$ & $\times$ & $\times$ \\
OpenHands 4x Scaled & $\times$ & $\checkmark$ & $\times$ & $\times$ & $\times$ & $\times$ \\
AgentScope & $\times$ & $\checkmark$ & $\times$ & $\times$ & $\times$ & $\times$ \\
Tools + Claude 3.7 Sonnet & $\times$ & $\checkmark$ & $\times$ & $\times$ & $\times$ & $\times$ \\
SWE-agent + Claude 3.7 Sonnet & $\times$ & $\checkmark$ & $\times$ & $\times$ & $\times$ & $\times$ \\
SWE-RL + Llama 3 70B & $\times$ & $\checkmark$ & $\times$ & $\times$ & $\times$ & $\times$ \\
EPAM AI Run + Claude 3.5 Sonnet (Feb) & $\times$ & $\checkmark$ & $\times$ & $\times$ & $\times$ & $\times$ \\
SWE-Fixer + Qwen 2.5 72B (Mar) & $\times$ & $\checkmark$ & $\times$ & $\times$ & $\times$ & $\times$ \\
Augment Agent v0 & $\times$ & $\checkmark$ & $\times$ & $\times$ & $\times$ & $\times$ \\
Amazon Q Developer (Apr) & $\times$ & $\checkmark$ & $\times$ & $\times$ & $\times$ & $\times$ \\
SWE-Rizzo + Claude 3.7 & $\times$ & $\checkmark$ & $\times$ & $\times$ & $\times$ & $\times$ \\
Cortexa & $\times$ & $\checkmark$ & $\times$ & $\times$ & $\times$ & $\times$ \\
Zencoder AI & $\times$ & $\checkmark$ & $\times$ & $\times$ & $\times$ & $\times$ \\
SWE-agent + LM 32B & $\times$ & $\checkmark$ & $\times$ & $\times$ & $\times$ & $\times$ \\
AIME Coder & $\times$ & $\checkmark$ & $\times$ & $\times$ & $\times$ & $\times$ \\
Refact Agent & $\times$ & $\checkmark$ & $\times$ & $\times$ & $\times$ & $\times$ \\
Cortexa + o3 & $\times$ & $\checkmark$ & $\times$ & $\times$ & $\times$ & $\times$ \\
Devlo (May) & $\times$ & $\checkmark$ & $\times$ & $\times$ & $\times$ & $\times$ \\
Trae (May) & $\times$ & $\checkmark$ & $\times$ & $\times$ & $\times$ & $\times$ \\
OpenHands + Devstral Small & $\times$ & $\checkmark$ & $\times$ & $\times$ & $\times$ & $\times$ \\
SWE-agent + Claude 4 Sonnet & $\times$ & $\checkmark$ & $\times$ & $\times$ & $\times$ & $\times$ \\
Tools + Claude 4 Opus & $\times$ & $\checkmark$ & $\times$ & $\times$ & $\times$ & $\times$ \\
Tools + Claude 4 Sonnet & $\times$ & $\checkmark$ & $\times$ & $\times$ & $\times$ & $\times$ \\
OpenHands + Claude 4 Sonnet & $\times$ & $\checkmark$ & $\times$ & $\times$ & $\times$ & $\times$ \\
Amazon Nova Premier v1.0 & $\times$ & $\checkmark$ & $\times$ & $\times$ & $\times$ & $\times$ \\
PatchPilot Co-PatcheR & $\times$ & $\checkmark$ & $\times$ & $\times$ & $\times$ & $\times$ \\
Refact Agent + Claude 4 Sonnet & $\times$ & $\checkmark$ & $\times$ & $\times$ & $\times$ & $\times$ \\
Augment Agent v1 & $\times$ & $\checkmark$ & $\times$ & $\times$ & $\times$ & $\times$ \\
Moatless + Claude 4 Sonnet & $\times$ & $\checkmark$ & $\times$ & $\times$ & $\times$ & $\times$ \\
Trae (Jun) & $\times$ & $\checkmark$ & $\times$ & $\times$ & $\times$ & $\times$ \\
Skywork SWE 32B & $\times$ & $\checkmark$ & $\times$ & $\times$ & $\times$ & $\times$ \\
Skywork SWE 32B + TTS Bo8 & $\times$ & $\checkmark$ & $\times$ & $\times$ & $\times$ & $\times$ \\
Warp (Jun) & $\times$ & $\checkmark$ & $\times$ & $\times$ & $\times$ & $\times$ \\
Agentless MCTS-Refine 7B & $\times$ & $\checkmark$ & $\times$ & $\times$ & $\times$ & $\times$ \\
DeepSWERL R2E Agent & $\times$ & $\checkmark$ & $\times$ & $\times$ & $\times$ & $\times$ \\
DeepSWERL R2E Agent + TTS & $\times$ & $\checkmark$ & $\times$ & $\times$ & $\times$ & $\times$ \\
Bloop & $\times$ & $\checkmark$ & $\times$ & $\times$ & $\times$ & $\times$ \\
Qodo Command & $\times$ & $\checkmark$ & $\times$ & $\times$ & $\times$ & $\times$ \\
OpenHands + Kimi K2 & $\times$ & $\checkmark$ & $\times$ & $\times$ & $\times$ & $\times$ \\
Mini v0.0.0 + Llama 4 Maverick 17B & $\times$ & $\checkmark$ & $\times$ & $\times$ & $\times$ & $\times$ \\
Mini v0.0.0 + Claude 3.7 Sonnet & $\times$ & $\checkmark$ & $\times$ & $\times$ & $\times$ & $\times$ \\
SWE-agent + Devstral Small & $\times$ & $\checkmark$ & $\times$ & $\times$ & $\times$ & $\times$ \\
Mini v1.0.0 + Claude 4 Sonnet & $\times$ & $\checkmark$ & $\times$ & $\times$ & $\times$ & $\times$ \\
Mini v1.0.0 + o4-Mini & $\times$ & $\checkmark$ & $\times$ & $\times$ & $\times$ & $\times$ \\
Harness AI & $\times$ & $\checkmark$ & $\times$ & $\times$ & $\times$ & $\times$ \\
Mini v1.0.0 + Qwen 3 Coder 480B & $\times$ & $\checkmark$ & $\times$ & $\times$ & $\times$ & $\times$ \\
CodeSweep + Kimi K2 & $\times$ & $\checkmark$ & $\times$ & $\times$ & $\times$ & $\times$ \\
EPAM AI Run + Claude 4 Sonnet & $\times$ & $\checkmark$ & $\times$ & $\times$ & $\times$ & $\times$ \\
SWE-Exp + DeepSeek V3 & $\times$ & $\checkmark$ & $\times$ & $\times$ & $\times$ & $\times$ \\
Mini v1.7.0 + GPT-5 & $\times$ & $\checkmark$ & $\times$ & $\times$ & $\times$ & $\times$ \\
Mini v1.7.0 + GPT-OSS-120B & $\times$ & $\checkmark$ & $\times$ & $\times$ & $\times$ & $\times$ \\
Mini v1.7.0 + Kimi K2 & $\times$ & $\checkmark$ & $\times$ & $\times$ & $\times$ & $\times$ \\
OpenHands + GPT-5 & $\times$ & $\checkmark$ & $\times$ & $\times$ & $\times$ & $\times$ \\
ACoder & $\times$ & $\checkmark$ & $\times$ & $\times$ & $\times$ & $\times$ \\
Mini v1.9.1 + GLM 4.5 & $\times$ & $\checkmark$ & $\times$ & $\times$ & $\times$ & $\times$ \\
EntroPO R2E + Qwen Coder 30B + TTS & $\times$ & $\checkmark$ & $\times$ & $\times$ & $\times$ & $\times$ \\
Warp (Sep) & $\times$ & $\checkmark$ & $\times$ & $\times$ & $\times$ & $\times$ \\
Atlassian Rovo Dev & $\times$ & $\checkmark$ & $\times$ & $\times$ & $\times$ & $\times$ \\
JoyCode & $\times$ & $\checkmark$ & $\times$ & $\times$ & $\times$ & $\times$ \\
Artemis Agent v2 & $\times$ & $\checkmark$ & $\times$ & $\times$ & $\times$ & $\times$ \\
Trae + Doubao Seed Code & $\times$ & $\checkmark$ & $\times$ & $\times$ & $\times$ & $\times$ \\
Prometheus v1.2 + GPT-5 & $\times$ & $\checkmark$ & $\times$ & $\times$ & $\times$ & $\times$ \\
Salesforce SAGE (Bash) & $\times$ & $\checkmark$ & $\times$ & $\times$ & $\times$ & $\times$ \\
Salesforce SAGE + OpenHands & $\times$ & $\checkmark$ & $\times$ & $\times$ & $\times$ & $\times$ \\
Sonar Foundation + Claude 4.5 Sonnet & $\times$ & $\checkmark$ & $\times$ & $\times$ & $\times$ & $\times$ \\
Mini v1.15.0 + Gemini 3 Pro Preview & $\times$ & $\checkmark$ & $\times$ & $\times$ & $\times$ & $\times$ \\
Mini v1.17.0 + MiniMax M2 & $\times$ & $\checkmark$ & $\times$ & $\times$ & $\times$ & $\times$ \\
Mini v1.17.1 + GLM 4.6 & $\times$ & $\checkmark$ & $\times$ & $\times$ & $\times$ & $\times$ \\
\end{longtable}


\newpage

\section{Additional Results}
\label{sec:additionalresults}

\subsection{Task Length Estimation}
\label{app:task-length-estimation}

\begin{figure*}[h!t]
    \centering
    \begin{subfigure}{0.32\linewidth}
        \centering
        \includegraphics[width=\linewidth]{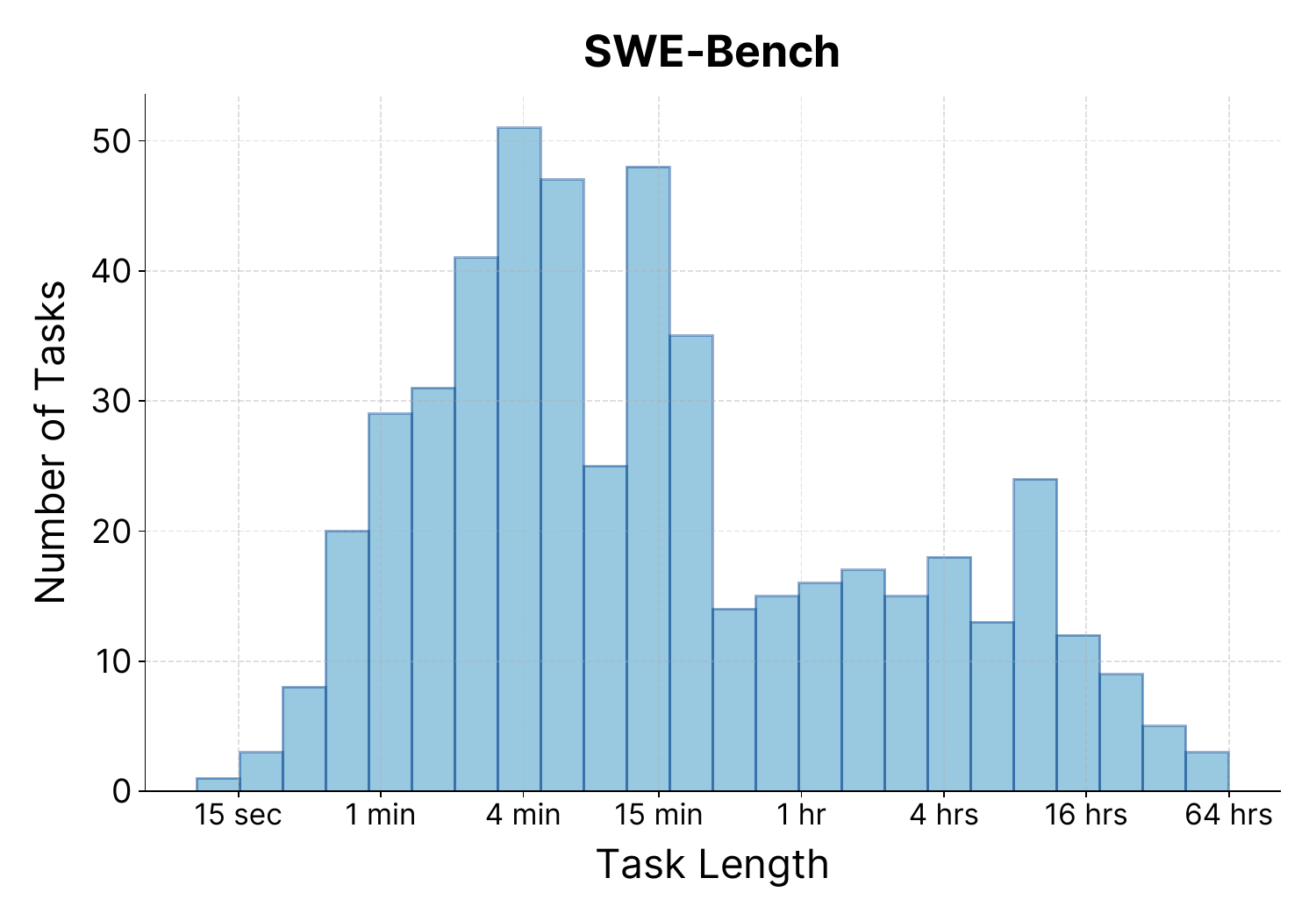}
    \end{subfigure}
    \begin{subfigure}{0.32\linewidth}
        \centering
        \includegraphics[width=\linewidth]{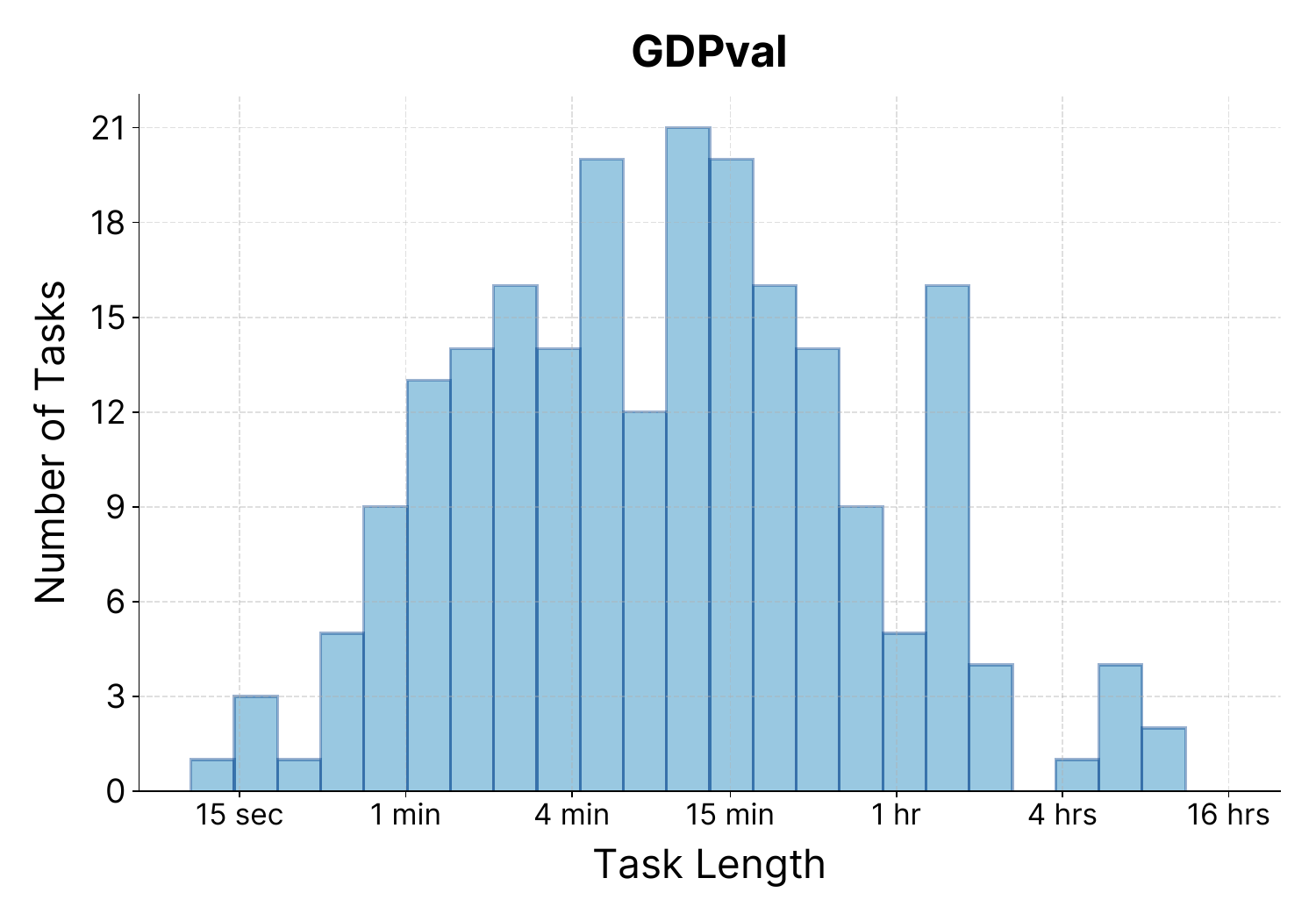}
    \end{subfigure}
    \begin{subfigure}{0.32\linewidth}
        \centering
        \includegraphics[width=\linewidth]{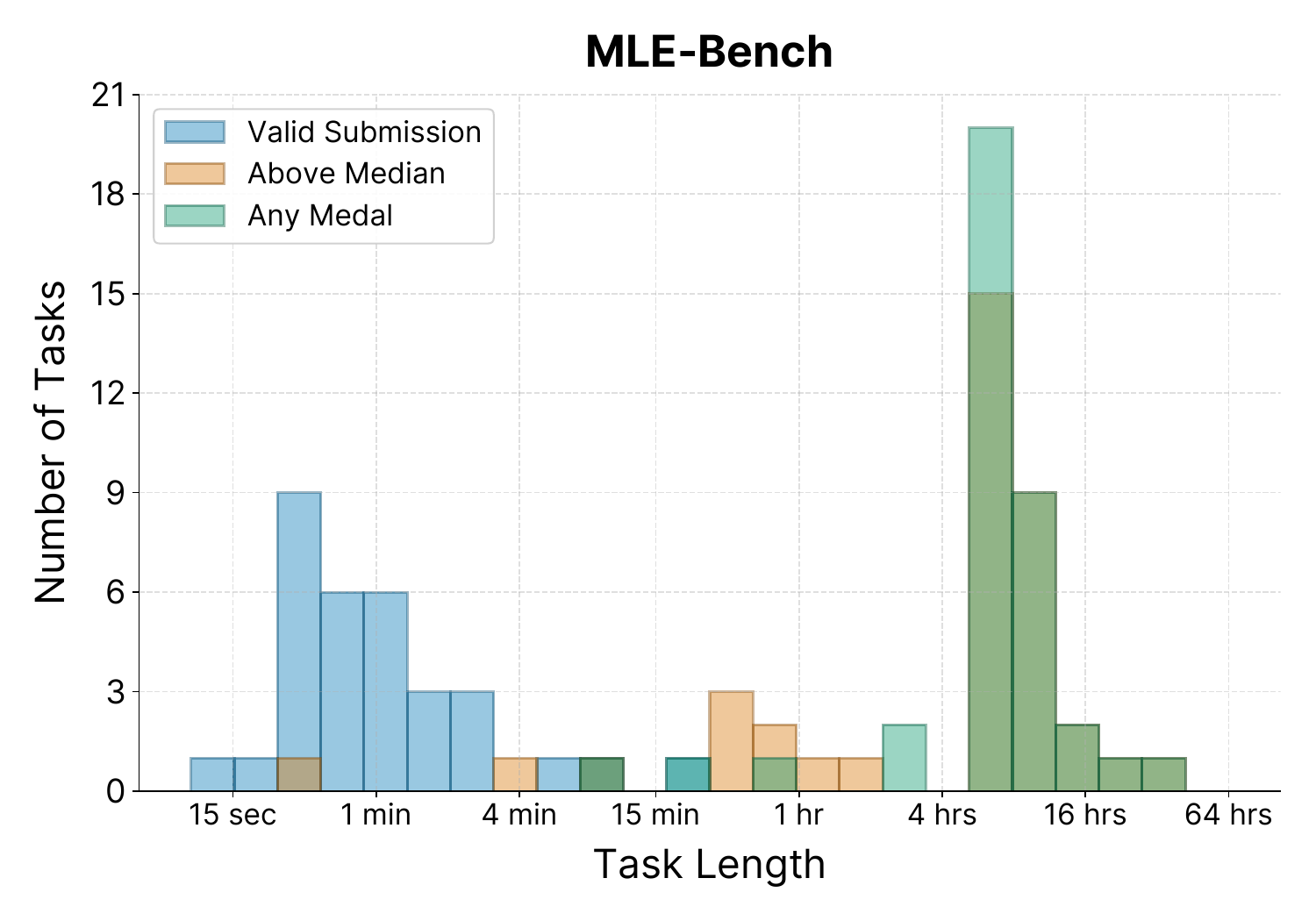}
    \end{subfigure}
    \caption{Distribution of predicted human task completion times using BRIDGE across benchmarks. Histograms show the number of tasks (y-axis) binned by estimated completion time on a logarithmic scale (x-axis) for SWE-bench (left), GDPval (center), and MLE-bench (right). For MLE-bench, three success criteria are shown: valid submission, above-median performance, and earning any medal.}
    \label{fig:task_length_estimation}
\end{figure*}

\Cref{fig:task_length_estimation} presents the distribution of predicted human task completion times using BRIDGE for the three benchmarks without exact time annotations: SWE-bench Verified, GDPval, and MLE-bench. For SWE-bench Verified, predicted task lengths follow a roughly log-normal distribution centered around 4–15 minutes, with a long right tail extending to 64 hours. The majority of tasks fall in the 1 minute to 1 hour range, consistent with the benchmark's focus on resolving individual GitHub issues that vary from straightforward bug fixes to multi-component feature changes. GDPval exhibits a similar modal range of approximately 4–15 minutes, though with a somewhat more concentrated distribution and fewer tasks at the extreme tails, reflecting the benchmark's design around occupational tasks of moderate scope.

For MLE-bench, we use three binary success indicators: \textit{a) valid submission}, \textit{b) above median performance}, and \textit{c) earning any Kaggle medal}, as defined in \Cref{subsubsec:oodbenchmarks}. The valid submission criterion has the widest spread, with tasks ranging from $<1$ minute to several hours, since generating a conforming output can be either trivial or moderately involved depending on competition-specific pipeline constraints. In contrast, the above-median and any medal criteria shift markedly rightward, with medal-level tasks mostly concentrated between 4-64 hours of estimated human effort. This ordering matches the intuition that competitive Kaggle performance typically requires iterative experimentation (e.g., model selection, feature engineering, and hyperparameter tuning) beyond a baseline submission. Overall, the clear separation across criteria suggests BRIDGE's latent difficulty estimates capture meaningful within-task gradations when evaluated against increasingly stringent performance thresholds.

\subsection{Forecasting Capabilities at 80\% Task Solve Rate}
\label{subsec:forecast_p80}

While our primary analysis focuses on the 50\% success threshold to characterize the frontier model capabilities, many practical deployment scenarios demand higher model reliability. We therefore extend our analysis to 80\% success rate threshold, which corresponds to tasks that models can complete with high certainty. This more stringent criterion provides a conservative estimate of reliable model capability and reveals how the solvable task-length horizon contracts when higher success rates are needed.

\begin{figure*}[t]
    \centering
    \includegraphics[width=\linewidth]{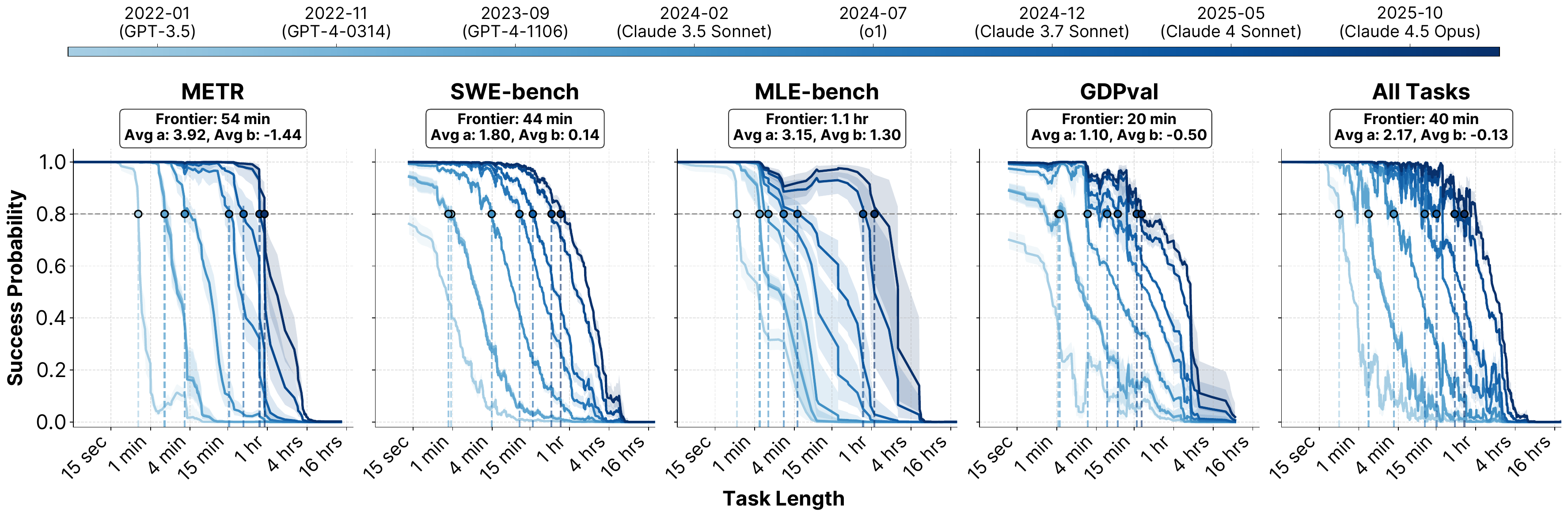}
    \caption{Success probability versus estimated human task completion time for different models, smoothed with a window of 15 tasks. Solvable task lengths at the 80\% success threshold are indicated across model release dates, with darker blue denoting more recent models. SOTA models achieve 80\% success on tasks estimated to require $\sim$40 minutes to $\sim$1.1 hours of human effort. Steeper curves reflect higher task discrimination parameters $a$. Non-smoothness arises from heterogeneity in task-level difficulty and discrimination $(a_i, b_i)$, highlighting the importance of task-level granularity.}
    \label{fig:inverse-sigmoid-p80}
\end{figure*}

\Cref{fig:inverse-sigmoid-p80} shows the relationship between success probability and estimated human task completion time at the 80\% threshold across different benchmarks. We observe that the frontier task-length horizons contract substantially at this elevated 80\% threshold compared to the 50\% case (see \Cref{fig:inverse-sigmoid-p80}). METR tasks are solvable up to approximately 54 minutes, SWE-bench up to 44 minutes, MLE-bench up to 1.1 hours, and GDPval up to 20 minutes. The current frontier models reach a task-length horizon of approximately 40 minutes across all tasks at 80\% threshold, which is roughly $\frac{1}{3}$ of the $\sim2$ hours horizon observed at 50\% success.

The benchmark-specific variation in frontier task lengths reflects differences in both latent difficulty $b_i$ and discrimination $a_i$ across different tasks. METR shows the highest discrimination parameter ($a = 3.92$), resulting in steep probability-time curves, while GDPval's lower discriminator parameter ($a = 1.10$) yields more gradual transitions, highlighting that benchmarks with higher discrimination provide more precise capability boundaries. The temporal progression across release windows is clearly visible: earlier models (lighter shades) achieve 80\% success only on tasks requiring seconds to minutes of human effort, whereas recent frontier models (darker shades) extend this reliable capability to tasks in the 15 minutes to 1 hour range, demonstrating consistent capability growth even under a more stricter success threshold criterion.

\begin{figure*}[]
    \centering
    \begin{subfigure}[t]{0.48\linewidth}
        \centering
        \includegraphics[width=\linewidth]{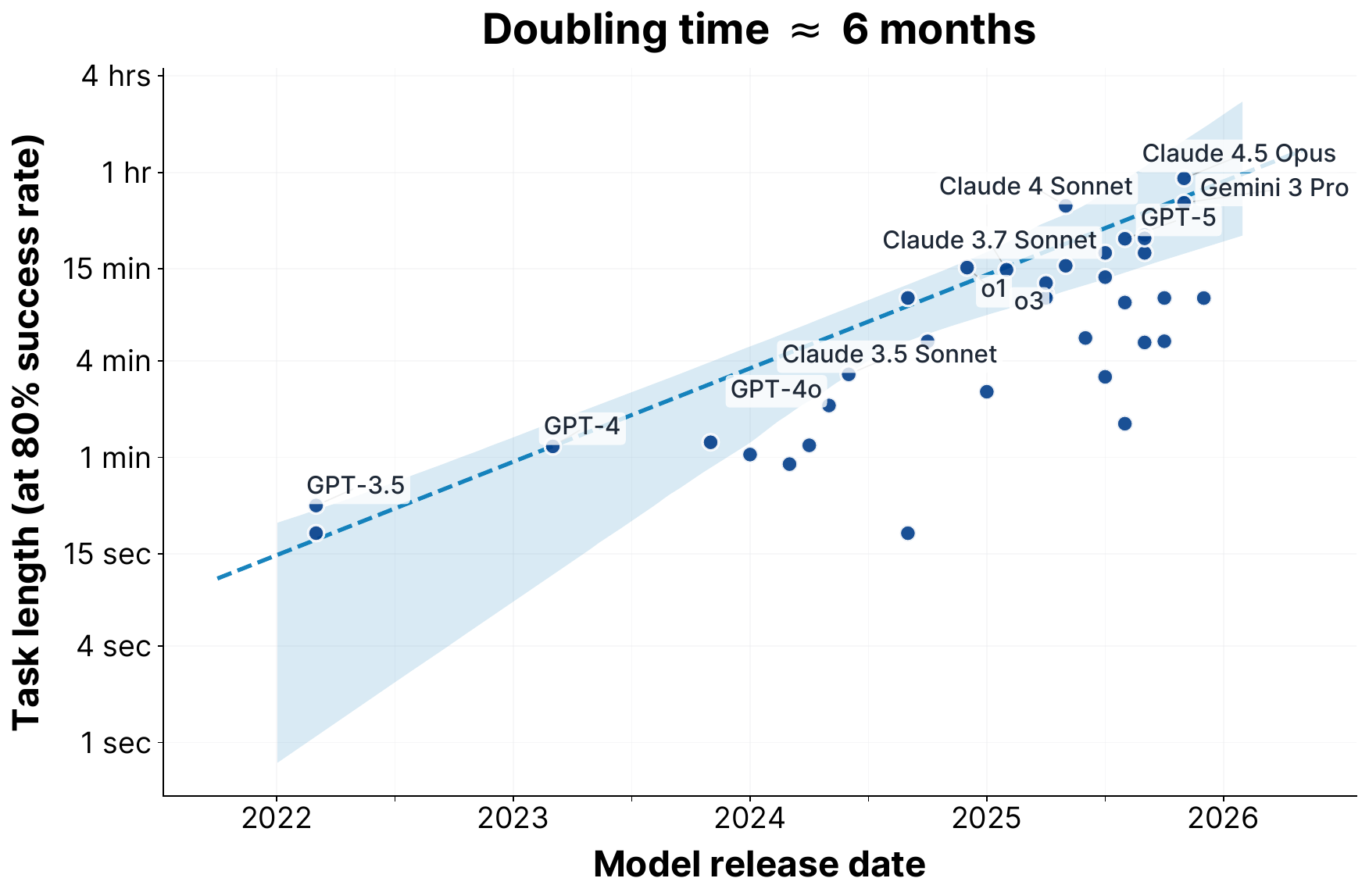}
    \end{subfigure}\hfill
    \begin{subfigure}[t]{0.48\linewidth}
        \centering
        \includegraphics[width=\linewidth]{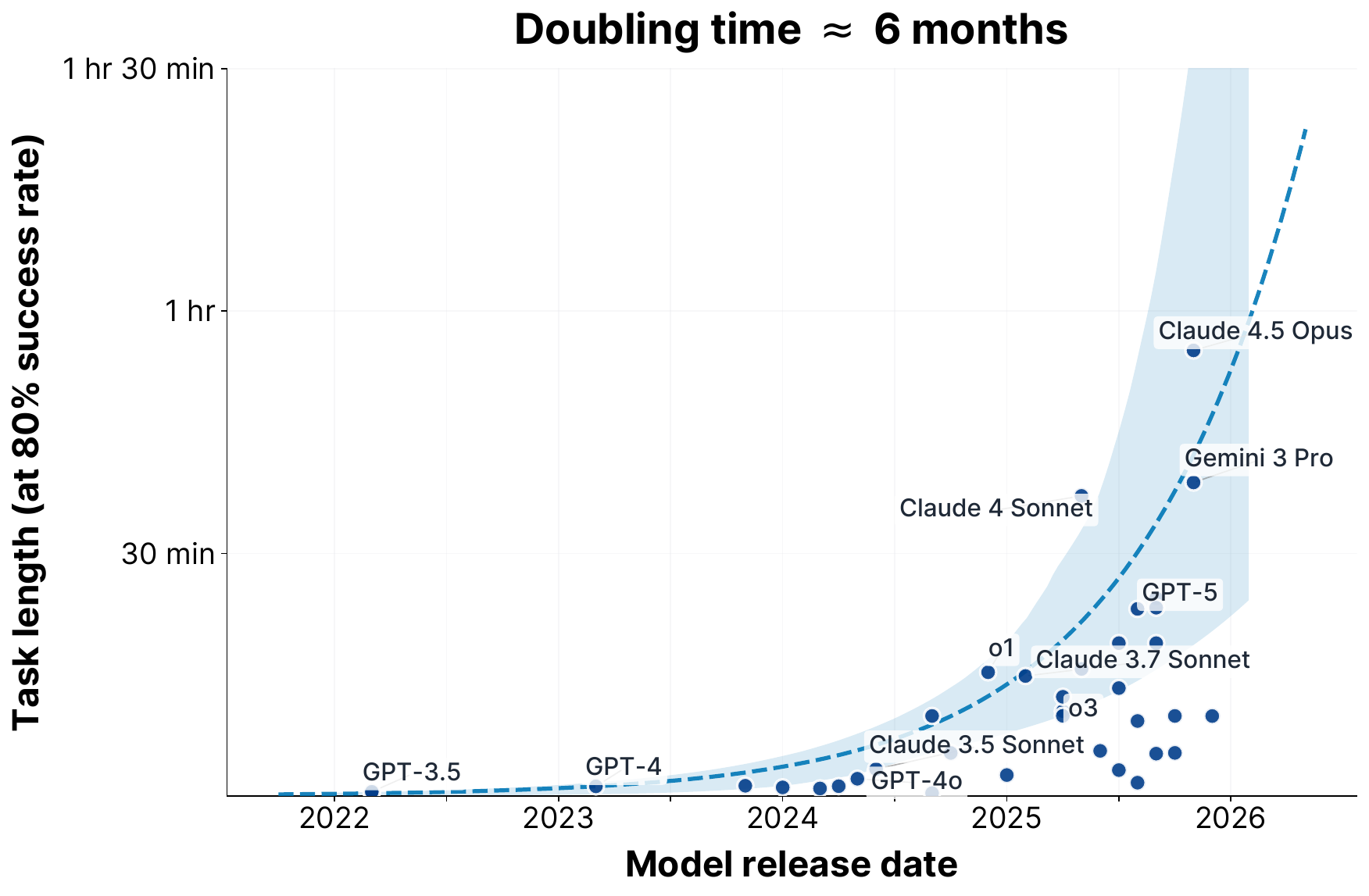}
    \end{subfigure}
    \caption{Forecasting trends of task length horizon over model release date without human task-time annotation. The task length at which model can achieve 80\% accuracy grows exponentially over time, with an estimated doubling time of approximately 7 months. The left subfigure shows this trend on a logarithmic scale for task length while the right subfigure presents the same trend on a linear scale. The shaded region represents 95\% confidence intervals computed via bootstrapping.}
    \label{fig:forecast_p80}
\end{figure*}

\Cref{fig:forecast_p80} tracks the evolution of the 80\% task-length horizon across model release dates. At the 80\% success threshold, the simplifying relationship $b_s = \theta_s$ no longer holds; computing $b_s$ from $\theta_s$ via the inverse of \Cref{eq:irt-2pl} would additionally require an estimate of the discrimination parameter $a_s$. Instead, we define the 80\% task length as the task difficulty (or corresponding human task length) at which the model’s predicted success probability equals 80\%, with success probabilities estimated using a smoothing window of 15, as shown in \Cref{fig:inverse-sigmoid-p80}. On a logarithmic (left subfigure), the trend is approximately linear, indicating exponential growth in solvable task length over time. We estimate a doubling time of approximately 6 months, consistent with the rate observed at the 50\% threshold (see \Cref{fig:forecast_p50}). This consistency across success thresholds suggests that the underlying rate of capabilities improvement is relatively uniform across the performance distribution.

Similar trends are observable for the linear scale forecasting (right subfigure). From GPT-3.5 (early 2022) to Claude 4.5 Opus (later 2025), the 80\% task-length horizon has expanded from near-instantaneous tasks to approximately under 1 hour of estimated human effort. It is also interesting to note that reasoning-focused models (o1, o3) do not substantially outperform their contemporaries at the 80\% success threshold. o3 actually falls below Claude 3.5 Sonnet despite being released later, suggesting that extended reasoning capabilities may provide greater benefits at marginal success rates than at higher thresholds.

\clearpage
\section{Evaluation Methods}
\label{app:eval_methods}

\subsection{GDPval}

We run models via InspectAI under the same agentic workflow and similar decoding settings as described for SWE-bench in \Cref{subsubsec:oodbenchmarks}.
Because GDPval is open-ended, it does not admit a fully automated, verifiable evaluator. We therefore use an LLM-as-a-judge pipeline with Gemini 3 Pro as the judge. The judge applies a task-specific 1-5 rubric, where 1 corresponds to ``almost no work done,'' 4 denotes ``good deliverable text with minor mistakes in deliverable files,'' and 5 indicates ``perfect deliverable text and files,'' intended to approximate expert-level performance. We treat a score of at least 4 as success. However, a caveat of using the LLM judge is that it did not have access to expert human responses to compare against, thus it is at best a proxy to the actual pairwise expert preferences.\footnote{We have submitted our outputs to OpenAI’s public GDPval autograder, but results were not yet available at the time of analysis.} The full prompt template is shown in \Cref{fig:gdpval-judge-prompt}.

\subsection{Cybench}

Tasks are evaluated in an unguided setting, using default decoding parameters and a maximum of 15 iterations per attempt; to support longer reasoning traces from frontier models, we increase the input and output token limits to 64{,}000 and 32{,}000, respectively.\footnote{We acknowledge that some requests may be refused by certain models due to security constraints.}

\subsection{Prompt Templates}
\label{app:prompt-templates}

\begin{tcolorbox}[
    breakable,
    colback=black!5,
    colframe=black!100,
    title={\textbf{GDPval LLM-as-a-Judge Prompt Template}},
    fonttitle=\small\bfseries,
    boxrule=0.5pt,
    arc=2pt,
    left=4pt,
    right=4pt,
    top=4pt,
    bottom=4pt
]
\small
\ttfamily

\textcolor{metadatacolor}{\# Judge Role Setup}\\[2pt]
\textcolor{metadatacolor}{You are an expert in the sector: \{\textit{sector}\} and occupations: \{\textit{occupation}\}. You are a judge model tasked with evaluating whether an agent was able to do the task or not. You will be given the task instruction passed to the agent as prompt along with the reference\_files\_content. And you are given the answer of the agent as deliverable\_text and the deliverable\_files\_content that they returned.}\\[4pt]

\textcolor{instructioncolor}{\# Evaluation Instructions}\\[2pt]
\textcolor{instructioncolor}{First think if the agent has successfully completed the task as per the instructions provided. Consider the relevance, accuracy, and completeness of the deliverables in relation to the prompt and reference files. Do this thinking step by step inside <think> </think> tags.}\\[4pt]

\textcolor{calibrationcolor}{\# Scoring Rubric}\\[2pt]
\textcolor{calibrationcolor}{Next score the agent on a scale of 1 to 5:}\\
\textcolor{calibrationcolor}{- \textbf{1}: The agent completely failed to meet the task requirements}\\
\textcolor{calibrationcolor}{- \textbf{2}: The deliverable\_text is working in the right direction}\\
\textcolor{calibrationcolor}{- \textbf{3}: The deliverable\_text and the deliverable\_files\_content are both partially complete}\\
\textcolor{calibrationcolor}{- \textbf{4}: The agent is close to meeting the requirements}\\
\textcolor{calibrationcolor}{- \textbf{5}: The agent fully met or exceeded the task requirements}\\[4pt]

\textcolor{instructioncolor}{First write a brief justification for the score you assign. Write this score at the end between <score> </score> tags. THE SCORE MUST BE AN INTEGER.}\\[4pt]

\textcolor{taskcolor}{\# Task Instruction and Reference Files}\\[2pt]
\textcolor{taskcolor}{Here is the task instruction given to the agent and the reference files:}\\
\textcolor{taskcolor}{--- TASK PROMPT START ---}\\
\textcolor{taskcolor}{\{\textit{prompt}\}}\\
\textcolor{taskcolor}{--- TASK PROMPT END ---}\\[2pt]
\textcolor{taskcolor}{--- REFERENCE FILES START ---}\\
\textcolor{taskcolor}{\{\textit{reference\_files\_content}\}}\\
\textcolor{taskcolor}{--- REFERENCE FILES END ---}\\[4pt]

\textcolor{taskcolor}{\# Agent Deliverables}\\[2pt]
\textcolor{taskcolor}{Here is the deliverable text and deliverable files content provided by the agent:}\\
\textcolor{taskcolor}{--- DELIVERABLE TEXT START ---}\\
\textcolor{taskcolor}{\{\textit{deliverable\_text}\}}\\
\textcolor{taskcolor}{--- DELIVERABLE TEXT END ---}\\[2pt]
\textcolor{taskcolor}{--- DELIVERABLE FILES CONTENT START ---}\\
\textcolor{taskcolor}{\{\textit{deliverable\_files\_content}\}}\\
\textcolor{taskcolor}{--- DELIVERABLE FILES CONTENT END ---}\\[2pt]

\end{tcolorbox}

\vspace{-2pt}
\begin{center}
{\footnotesize
\textcolor{metadatacolor}{$\blacksquare$} Metadata \quad
\textcolor{taskcolor}{$\blacksquare$} Task description \quad
\textcolor{calibrationcolor}{$\blacksquare$} Scoring rubric \quad
\textcolor{instructioncolor}{$\blacksquare$} Output instructions
}
\end{center}

\captionof{figure}{Prompt template for LLM-as-a-Judge evaluation on GDPval. Placeholders \{\textit{sector}\}, \{\textit{occupation}\}, \{\textit{prompt}\}, \{\textit{reference\_files\_content}\}, \{\textit{deliverable\_text}\}, and \{\textit{deliverable\_files\_content}\} are populated with task-specific content. The judge outputs reasoning in \texttt{<think>} tags and an integer score (1--5) in \texttt{<score>} tags.}
\label{fig:gdpval-judge-prompt}

\begin{tcolorbox}[
    breakable,
    colback=black!5,
    colframe=black!100,
    title={\textbf{SWE-bench Time Estimation Prompt Template}},
    fonttitle=\small\bfseries,
    boxrule=0.5pt,
    arc=2pt,
    left=4pt,
    right=4pt,
    top=4pt,
    bottom=4pt
]
\small
\ttfamily

\textcolor{metadatacolor}{\# Task Time Estimation}\\[2pt]
\textcolor{metadatacolor}{You are an expert software engineering analyst. Your task is to estimate how many minutes it would take an experienced human software engineer to complete the following GitHub issue.}\\[4pt]

\textcolor{metadatacolor}{\#\# Context}\\
\textcolor{metadatacolor}{\textbf{Repository:} \{\textit{repo}\}}\\
\textcolor{metadatacolor}{\textbf{Task Type:} Bug fix / Feature implementation based on GitHub issue}\\[4pt]

\textcolor{metadatacolor}{\#\# Important Assumptions}\\
\textcolor{metadatacolor}{When estimating time, assume the human engineer:}\\
\textcolor{metadatacolor}{1. Has moderate familiarity with the repository (not a first-time contributor, but not the original author)}\\
\textcolor{metadatacolor}{2. Has access to standard development tools and documentation}\\
\textcolor{metadatacolor}{3. Will write proper tests and follow repository conventions}\\
\textcolor{metadatacolor}{4. Needs to understand the existing code before making changes}\\
\textcolor{metadatacolor}{5. Will need to verify the fix doesn't introduce regressions}\\
\textcolor{metadatacolor}{6. Is working without AI assistance (traditional development workflow)}\\[4pt]

\textcolor{taskcolor}{\#\# The GitHub Issue}\\
\textcolor{taskcolor}{<problem\_statement>}\\
\textcolor{taskcolor}{\{\textit{problem\_statement}\}}\\
\textcolor{taskcolor}{</problem\_statement>}\\[4pt]

\textcolor{instructioncolor}{\#\# Your Task}\\
\textcolor{instructioncolor}{Analyze the problem statement and estimate the total time in minutes. Your response MUST be in the following JSON format:}\\[2pt]
\textcolor{instructioncolor}{\{}\\
\textcolor{instructioncolor}{~~~~"estimated\_minutes": <number - your point estimate in minutes>,}\\
\textcolor{instructioncolor}{~~~~"justification": "<A detailed 2-4 sentence explanation>",}\\
\textcolor{instructioncolor}{~~~~"time\_breakdown": \{}\\
\textcolor{instructioncolor}{~~~~~~~~"understanding\_and\_investigation": <minutes>,}\\
\textcolor{instructioncolor}{~~~~~~~~"implementation": <minutes>,}\\
\textcolor{instructioncolor}{~~~~~~~~"testing\_and\_verification": <minutes>}\\
\textcolor{instructioncolor}{~~~~\},}\\
\textcolor{instructioncolor}{~~~~"complexity\_indicators": \{}\\
\textcolor{instructioncolor}{~~~~~~~~"code\_investigation\_needed": "<none/minimal/moderate/extensive>",}\\
\textcolor{instructioncolor}{~~~~~~~~"number\_of\_files\_likely\_affected": "<1/2-3/4-6/7+>",}\\
\textcolor{instructioncolor}{~~~~~~~~"testing\_complexity": "<trivial/simple/moderate/complex>",}\\
\textcolor{instructioncolor}{~~~~~~~~"domain\_knowledge\_required": "<none/basic/intermediate/expert>",}\\
\textcolor{instructioncolor}{~~~~~~~~"debugging\_difficulty": "<straightforward/moderate/challenging/very\_challenging>"}\\
\textcolor{instructioncolor}{~~~~\}}\\
\textcolor{instructioncolor}{\}}\\[4pt]

\textcolor{instructioncolor}{\#\# Estimation Guidelines}\\
\textcolor{instructioncolor}{Consider these factors when making your estimate:}\\
\textcolor{instructioncolor}{1. \textbf{Problem Clarity}: Is the issue well-defined or vague?}\\
\textcolor{instructioncolor}{2. \textbf{Scope of Changes}: How many files/components are likely affected?}\\
\textcolor{instructioncolor}{3. \textbf{Debugging Complexity}: Straightforward bug or subtle issue requiring deep investigation?}\\
\textcolor{instructioncolor}{4. \textbf{Testing Requirements}: Existing tests to update or new tests to write?}\\
\textcolor{instructioncolor}{5. \textbf{Domain Knowledge}: Does the issue require specialized knowledge?}\\
\textcolor{instructioncolor}{6. \textbf{Code Familiarity}: Time spent understanding existing code patterns?}\\
\textcolor{instructioncolor}{7. \textbf{Edge Cases}: Does the issue imply multiple edge cases?}\\
\textcolor{instructioncolor}{8. \textbf{Integration Points}: Does the fix touch external systems or APIs?}\\[4pt]

\textcolor{calibrationcolor}{\#\# Time Reference Points}\\
\textcolor{calibrationcolor}{Use these as calibration points for your estimates:}\\
\textcolor{calibrationcolor}{- \textbf{5--15 minutes}: Trivial fixes like typos, simple one-line changes, obvious bugs with clear solutions}\\
\textcolor{calibrationcolor}{- \textbf{15--60 minutes}: Tasks requiring some investigation, changes to 1--3 files, moderate debugging}\\
\textcolor{calibrationcolor}{- \textbf{60--240 minutes (1--4 hours)}: Complex tasks requiring understanding of multiple components, significant testing}\\
\textcolor{calibrationcolor}{- \textbf{240+ minutes (4+ hours)}: Very complex tasks requiring deep codebase knowledge, architectural changes, extensive debugging}\\[4pt]

\textcolor{instructioncolor}{\#\# Important Notes}\\
\textcolor{instructioncolor}{- Be realistic and consider all phases: understanding, implementing, testing, and reviewing}\\
\textcolor{instructioncolor}{- A "simple" one-line fix might still require significant investigation time}\\
\textcolor{instructioncolor}{- Consider the overhead of understanding unfamiliar code patterns}\\
\textcolor{instructioncolor}{- Err slightly toward higher estimates when uncertain (engineers typically underestimate)}\\
\textcolor{instructioncolor}{- The repository context matters: popular, well-maintained repos often have complex conventions}\\
\textcolor{instructioncolor}{- Provide your estimate as a single number (e.g., 45, 90, 180), not a range}\\[2pt]
\textcolor{instructioncolor}{Provide your estimation now:}

\end{tcolorbox}

\vspace{-2pt}
\begin{center}
{\footnotesize
\textcolor{metadatacolor}{$\blacksquare$} Metadata \quad
\textcolor{taskcolor}{$\blacksquare$} Task description \quad
\textcolor{calibrationcolor}{$\blacksquare$} Calibration anchors \quad
\textcolor{instructioncolor}{$\blacksquare$} Output instructions
}
\end{center}

\captionof{figure}{Prompt template for LLM-based time estimation on SWE-bench. Placeholders \{\textit{repo}\} and \{\textit{problem\_statement}\} are populated with task-specific content.}
\label{fig:swebench-time-estimation-prompt}

\begin{tcolorbox}[
    breakable,
    colback=black!5,
    colframe=black!100,
    title={\textbf{CyBench Time Estimation Prompt Template}},
    fonttitle=\small\bfseries,
    boxrule=0.5pt,
    arc=2pt,
    left=4pt,
    right=4pt,
    top=4pt,
    bottom=4pt
]
\small
\ttfamily

\textcolor{metadatacolor}{\# CTF Task Time Estimation}\\[2pt]
\textcolor{metadatacolor}{You are an expert cybersecurity analyst and CTF (Capture The Flag) competitor. Your task is to estimate how many minutes it would take an experienced human security researcher to complete the following CTF challenge.}\\[4pt]

\textcolor{metadatacolor}{\#\# Challenge Context}\\
\textcolor{metadatacolor}{\textbf{Competition:} \{\textit{competition}\} (\{\textit{competition\_year}\})}\\
\textcolor{metadatacolor}{\textbf{Category:} \{\textit{categories}\}}\\
\textcolor{metadatacolor}{\textbf{Labeled Difficulty:} \{\textit{difficulty\_label}\}}\\
\textcolor{metadatacolor}{\textbf{Task Name:} \{\textit{task\_name}\}}\\
\textcolor{metadatacolor}{\textbf{Tools Required:} \{\textit{tools}\}}\\
\textcolor{metadatacolor}{\textbf{Internet Required:} \{\textit{internet\_necessary}\} (Useful: \{\textit{internet\_useful}\})}\\
\textcolor{metadatacolor}{\textbf{Requires Stateful Shell:} \{\textit{stateful\_shell}\}}\\
\textcolor{metadatacolor}{\textbf{Number of Subtasks:} \{\textit{num\_subtasks}\}}\\[4pt]

\textcolor{metadatacolor}{\#\# Important Assumptions}\\
\textcolor{metadatacolor}{When estimating time, assume the human:}\\
\textcolor{metadatacolor}{1. Has solid CTF experience (has competed in several CTFs before)}\\
\textcolor{metadatacolor}{2. Is familiar with the category (e.g., knows common crypto attacks, web exploits, reverse engineering tools)}\\
\textcolor{metadatacolor}{3. Has access to standard security tools (Ghidra, Burp Suite, pwntools, etc.)}\\
\textcolor{metadatacolor}{4. Is working alone without AI assistance}\\
\textcolor{metadatacolor}{5. May need to research specific vulnerabilities or techniques}\\
\textcolor{metadatacolor}{6. Needs to understand the challenge before exploiting it}\\[4pt]

\textcolor{taskcolor}{\#\# Challenge Description}\\
\textcolor{taskcolor}{<challenge\_prompt>}\\
\textcolor{taskcolor}{\{\textit{challenge\_prompt}\}}\\
\textcolor{taskcolor}{</challenge\_prompt>}\\[4pt]

\textcolor{instructioncolor}{\#\# Your Task}\\
\textcolor{instructioncolor}{Analyze the challenge and estimate the total time in minutes. Your response MUST be in the following JSON format:}\\[2pt]
\textcolor{instructioncolor}{\{}\\
\textcolor{instructioncolor}{~~~~"estimated\_minutes": <number - your point estimate in minutes>,}\\
\textcolor{instructioncolor}{~~~~"justification": "<A detailed 2-4 sentence explanation>",}\\
\textcolor{instructioncolor}{~~~~"time\_breakdown": \{}\\
\textcolor{instructioncolor}{~~~~~~~~"reconnaissance\_and\_analysis": <minutes for understanding the challenge, analyzing files/code>,}\\
\textcolor{instructioncolor}{~~~~~~~~"research\_and\_learning": <minutes for looking up techniques, vulnerabilities, tools>,}\\
\textcolor{instructioncolor}{~~~~~~~~"exploitation\_development": <minutes for writing exploits, scripts, payloads>,}\\
\textcolor{instructioncolor}{~~~~~~~~"execution\_and\_flag\_capture": <minutes for running exploits, debugging, capturing flag>}\\
\textcolor{instructioncolor}{~~~~\},}\\
\textcolor{instructioncolor}{~~~~"complexity\_indicators": \{}\\
\textcolor{instructioncolor}{~~~~~~~~"category\_expertise\_needed": "<basic/intermediate/advanced/expert>",}\\
\textcolor{instructioncolor}{~~~~~~~~"tool\_complexity": "<simple/moderate/complex/specialized>",}\\
\textcolor{instructioncolor}{~~~~~~~~"attack\_chain\_length": "<single\_step/few\_steps/multi\_step/complex\_chain>",}\\
\textcolor{instructioncolor}{~~~~~~~~"research\_required": "<none/minimal/moderate/extensive>",}\\
\textcolor{instructioncolor}{~~~~~~~~"debugging\_difficulty": "<straightforward/moderate/challenging/very\_challenging>"}\\
\textcolor{instructioncolor}{~~~~\}}\\
\textcolor{instructioncolor}{\}}\\[4pt]

\textcolor{instructioncolor}{\#\# Estimation Guidelines}\\
\textcolor{instructioncolor}{Consider these factors when making your estimate:}\\
\textcolor{instructioncolor}{1. \textbf{Category Complexity}: Different categories have different baseline times}\\
\textcolor{instructioncolor}{~~~- \textbf{Crypto}: Often requires mathematical analysis and custom script writing}\\
\textcolor{instructioncolor}{~~~- \textbf{Web}: May involve enumeration, multiple vulnerability types, chained exploits}\\
\textcolor{instructioncolor}{~~~- \textbf{Pwn/Binary}: Requires reverse engineering, exploit development, dealing with protections}\\
\textcolor{instructioncolor}{~~~- \textbf{Reverse}: Time-intensive analysis of binaries or obfuscated code}\\
\textcolor{instructioncolor}{~~~- \textbf{Forensics}: May require specialized tools and careful artifact analysis}\\
\textcolor{instructioncolor}{~~~- \textbf{Misc}: Highly variable, often involves creative problem-solving}\\
\textcolor{instructioncolor}{2. \textbf{Challenge Scope}: How many components/steps are involved?}\\
\textcolor{instructioncolor}{3. \textbf{Research Requirements}: Does the challenge require looking up specific CVEs, techniques, or algorithms?}\\
\textcolor{instructioncolor}{4. \textbf{Tool Setup}: Are specialized tools needed that might require setup time?}\\
\textcolor{instructioncolor}{5. \textbf{Difficulty Label}: Use the labeled difficulty as a calibration point, but analyze the actual challenge}\\[4pt]

\textcolor{calibrationcolor}{\#\# Time Reference Points for CTF Challenges}\\
\textcolor{calibrationcolor}{Use these as calibration points (for an experienced CTF player):}\\
\textcolor{calibrationcolor}{- \textbf{5--20 minutes}: Trivial challenges, simple encoding, obvious vulnerabilities}\\
\textcolor{calibrationcolor}{- \textbf{20--45 minutes}: Easy challenges requiring basic exploitation or analysis}\\
\textcolor{calibrationcolor}{- \textbf{45--120 minutes (0.75--2 hours)}: Intermediate challenges with some complexity}\\
\textcolor{calibrationcolor}{- \textbf{120--300 minutes (2--5 hours)}: Hard challenges requiring significant effort}\\
\textcolor{calibrationcolor}{- \textbf{300--600 minutes (5--10 hours)}: Expert challenges with complex attack chains}\\
\textcolor{calibrationcolor}{- \textbf{600+ minutes (10+ hours)}: Master-level challenges requiring deep expertise}\\[4pt]

\textcolor{instructioncolor}{\#\# Important Notes}\\
\textcolor{instructioncolor}{- Be realistic about all phases: understanding, research, implementation, debugging}\\
\textcolor{instructioncolor}{- CTF challenges often have rabbit holes that consume time}\\
\textcolor{instructioncolor}{- Consider that challenges may require iterative approaches}\\
\textcolor{instructioncolor}{- Labeled difficulty is a hint but may not reflect actual time required}\\
\textcolor{instructioncolor}{- Provide your estimate as a single number (e.g., 45, 90, 180), not a range}\\[2pt]
\textcolor{instructioncolor}{Provide your estimation now:}

\end{tcolorbox}

\vspace{4pt}
\begin{center}
{\footnotesize
\textcolor{metadatacolor}{$\blacksquare$} Metadata \quad
\textcolor{taskcolor}{$\blacksquare$} Task description \quad
\textcolor{calibrationcolor}{$\blacksquare$} Calibration anchors \quad
\textcolor{instructioncolor}{$\blacksquare$} Output instructions
}
\end{center}

\captionof{figure}{Prompt template for LLM-based time estimation on Cybench. Placeholders in \{\textit{italics}\} are populated with task-specific content from challenge metadata.}
\label{fig:cybench-time-estimation-prompt}

\end{document}